\documentclass[10pt,twocolumn,letterpaper]{article}

\usepackage{cvpr}              

\usepackage{graphicx}
\usepackage{amsmath}
\usepackage{amssymb}
\usepackage{booktabs}

\usepackage{times}
\usepackage{epsfig}
\usepackage{graphicx}
\usepackage{amsmath}
\usepackage{amssymb}
\usepackage{booktabs}
\usepackage{xr}
\usepackage[ruled,vlined]{algorithm2e}
\SetKwInput{KwInput}{Input}
\SetKwInput{KwOutput}{Output}
\DeclareMathOperator*{\argmax}{arg\,max}
\DeclareMathOperator*{\argmin}{arg\,min}
\usepackage{xcolor}
\newcommand{\red}[1]{\textcolor{red}{#1}}
\newcommand{\blue}[1]{\textcolor{blue}{#1}}
\newcommand{\green}[1]{\textcolor{green}{#1}}

\newcommand{\orange}[1]{\textcolor{orange}{#1}}
\newcommand{\pink}[1]{\textcolor{pink}{#1}}
\newcommand{\purple}[1]{\textcolor{purple}{#1}}
\newcommand{\gray}[1]{\textcolor{gray}{#1}}
\newcommand{\olive}[1]{\textcolor{olive}{#1}}

\newcommand{\PatchCore}{\textit{PatchCore}~}
\newcommand{\PatchCoreN}[1]{\textit{PatchCore}$-{#1}\%$}

\usepackage{multicol}

\makeatletter
\newcommand{\maketitlepage}{%
    \let\thanks\@gobble
    \let\footnote\@gobble
    \if@twocolumn
      \ifnum \col@number=\@ne
        \@maketitle
      \else
        \twocolumn[\@maketitle]%
      \fi
    \else
      \@maketitle
    \fi
    \thispagestyle{empty}
}
\makeatother

\usepackage{subcaption}

\usepackage[pagebackref,breaklinks,colorlinks]{hyperref}

\usepackage[capitalize]{cleveref}
\crefname{section}{Sec.}{Secs.}
\Crefname{section}{Section}{Sections}
\Crefname{table}{Table}{Tables}
\crefname{table}{Tab.}{Tabs.}

\def\cvprPaperID{7097} 
\def\confName{CVPR}
\def\confYear{2022}

\begin{document}

\title{Towards Total Recall in Industrial Anomaly Detection}

\author{Karsten Roth$^{1,*}$, Latha Pemula$^{2}$, Joaquin Zepeda$^{2}$, Bernhard Schölkopf$^{2}$, Thomas Brox$^{2}$, Peter Gehler$^{2}$\\
$^1$University of Tübingen\hspace{6pt} $^2$Amazon AWS\hspace{6pt}
}

\let\svthefootnote\thefootnote
\newcommand\freefootnote[1]{%
  \let\thefootnote\relax%
  \footnotetext{#1}%
  \let\thefootnote\svthefootnote%
}

\maketitle

\begin{abstract}
Being able to spot defective parts is a critical component in large-scale industrial manufacturing. A particular challenge that we address in this work is the cold-start problem:
fit a model using nominal (non-defective) example images only.
While handcrafted solutions per class are possible, the goal is to build systems that work well simultaneously on many different tasks automatically. The best peforming approaches
combine embeddings from ImageNet models with an outlier detection model.
In this paper, we extend on this line of work and propose \textbf{PatchCore}, which uses a maximally representative memory bank of nominal patch-features.
PatchCore offers competitive inference times while achieving state-of-the-art performance for both detection and localization. On the challenging, widely used MVTec AD benchmark 
PatchCore achieves an image-level anomaly detection AUROC score of up to $99.6\%$, more than halving the error compared to the next best competitor. We further report
competitive results on two additional datasets and also find competitive results in the few samples regime.\freefootnote{$^*$ Work done during a research internship at Amazon AWS.} Code: \href{github.com/amazon-research/patchcore-inspection}{github.com/amazon-research/patchcore-inspection}.
\end{abstract}

\section{Introduction}\label{sec:introduction}

\begin{figure}
\centering
\includegraphics[width=0.48\textwidth]{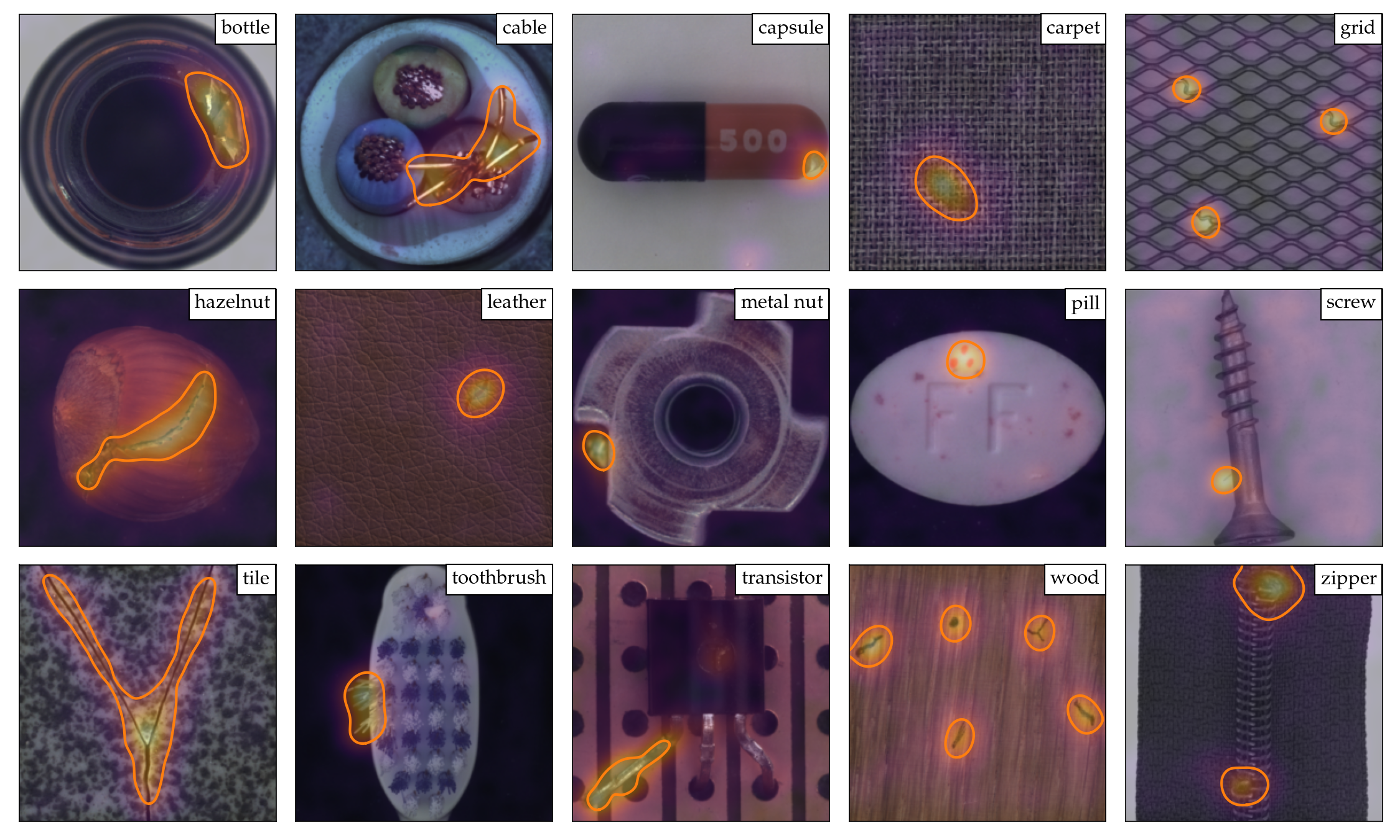}
\vspace{-3pt}
\caption{Examples from the MVTec benchmark datasets. Superimposed on the images are the segmentation results from \textit{PatchCore}. The \orange{orange} boundary denotes anomaly contours of actual segmentation maps for anomalies such as broken glass, scratches, burns or structural changes in \blue{blue}-\orange{orange} color gradients.}
\label{fig:segmentation_examples}
\vspace{-3pt}
\end{figure}

The ability to detect unusual patterns in images is a feature deeply ingrained in human cognition. Humans can differentiate between expected variance in the data and outliers after having only seen a small number of normal instances.
In this work we address the computational version of this problem, \emph{cold-start}\footnote{Commonly also dubbed one-class classification (OCC).} anomaly detection for visual inspection of industrial image data. It arises in many industrial scenarios
where it is easy to acquire imagery of normal examples but costly and complicated to specify the expected defect variations in full.
This task is naturally cast as a out-of-distribution detection problem where a model needs
to distinguish between samples being drawn from the training data distribution and those outside its support. Industrial visual defect classification is especially hard, as errors can vary from subtle changes such as thin scratches to larger structural defects like missing components \cite{mvtec}. Some examples from the MVTec AD benchmark along with results from our proposed method are shown in Figure~\ref{fig:segmentation_examples}.
Existing work on cold-start, industrial visual anomaly detection relies on learning a model of the nominal distribution via auto-encoding methods \cite{sakurada2014autoencoder,nguyen2019anomalyhypothesis,davletshina2020unsupervised}, GANs \cite{akcay2018ganomaly,pidhorskyi2018generative,sabokrou2018oneclass}, or other unsupervised adaptation methods \cite{yi2020patch,rudolph2021differnet}.
Recently, \cite{bergman2020deep,cohen2020transformerbased} proposed to leverag common deep representations from ImageNet classification without adaptation to the target distribution.
Despite the missing adaptation, these models offer strong anomaly detection performance and even solid spatial localization of the defects.
The key principle behind these techniques is a feature matching between the test sample and the nominal samples while exploiting the multi-scale nature of deep feature representations.
Subtle, fine-grained defect segmentation is covered by high-resolution features, whereas structural deviations and full image-level anomaly detection are supposed to be covered by features at much higher abstraction levels.
The inherent downside of this approach, since it is non-adaptive, is the limited matching confidence at the higher abstraction levels: high-level abstract features from ImageNet training coincide little with the abstract features required in an industrial environment. In addition, nominal context usable by these methods at test time is effectively limited by the small number of extractable high-level feature representations. 

In this paper, we present \PatchCore as an effective remedy by (1) maximizing nominal information available at test time, (2) reducing biases towards ImageNet classes and (3) retaining high inference speeds. 
Relying on the fact that an image can be already classified as anomalous as soon as a single patch is anomalous \cite{yi2020patch,defard2020padim}, \PatchCore achieves this by utilizing locally aggregated, mid-level features patches. The usage of mid-level network patch features allows \PatchCore to operate with minimal bias towards ImageNet classes on a high resolution, while a feature aggregation over a local neighbourhood ensures retention of sufficient spatial context. This results in an extensive memory bank allowing \PatchCore to optimally leverage available nominal context at test time.
Finally, for practical applicability, \PatchCore additionally introduces greedy coreset subsampling \cite{agarwal2004coresetgeometry} for nominal feature banks as a key element to both reduce redundancy in the extracted, patch-level memory bank as well as significantly bringing down storage memory and inference time, making \PatchCore very attractive for realistic industrial use cases.

Thorough experiments on the diverse MVTec AD \cite{mvtec} as well as the specialized Magnetic Tile Defects (MTD)~\cite{magnetictiles} industrial anomaly detection benchmarks showcase the power of \PatchCore for industrial anomaly detection. 
It achieves state-of-the-art image-level detection scores on MVTec AD and MTD, with nearly perfect scores on MVTec AD (up to AUROC $99.6\%$), reducing detection error of previous methods by \textbf{more than half}, as well as state-of-the-art industrial anomaly localization performance. \PatchCore achieves this while retaining fast inference times without requiring training on the dataset at hand. This makes \textit{PatchCore} very attractive for practical use in industrial anomaly detection.
In addition, further experiments showcase the high sample efficiency of \textit{PatchCore}, matching existing anomaly detection methods in performance while using only a fraction of the nominal training data.



\section{Related Works}\label{sec:related_works}
Most anomaly detection models rely on the ability to learn representations inherent to the nominal data.
This can be achieved for example through the usage of autoencoding models \cite{sakurada2014autoencoder}. To encourage better estimation of the nominal feature distribution, extensions based on Gaussian mixture models \cite{zong2018deep}, generative adversarial training objectives \cite{pidhorskyi2018generative,akcay2018ganomaly,sabokrou2018oneclass}, invariance towards predefined physical augmentations \cite{huang2019inversetransform}, robustness of hidden features to reintroduction of reconstructions \cite{kim2020rapp}, prototypical memory banks \cite{gong2019memory}, attention-guidance \cite{venkataramanan2019cavga}, structural objectives \cite{wang2004ssim,bergmann2019ssim} or constrained representation spaces \cite{perera2019ocgan} have been proposed.
Other unsupervised representation learning methods can similarly be utilised, such as via GANs  \cite{deecke2018anomaly}, learning to predict predefined geometric transformations \cite{izhak2018geometric} or via normalizing flows \cite{rudolph2021differnet}.
Given respective nominal representations and novel test representations, anomaly detection can then be a simple matter of reconstruction errors \cite{sakurada2014autoencoder}, distances to $k$ nearest neighbours \cite{eskin2002} or finetuning of a one-class classification model such as OC-SVMs \cite{schoelkopf2000ocsvm} or SVDD \cite{tax2004svdd,yi2020patch} on top of these features.
For the majority of these approaches, anomaly localization comes naturally based on pixel-wise reconstruction errors, saliency-based approaches such as GradCAM \cite{gradcam} or XRAI \cite{xrai} can be used for anomaly segmentation \cite{venkataramanan2019cavga,rudolph2021differnet,salehi2020multiresolution} as well.

\textbf{Industrial Anomaly Detection.} While literature on general anomaly detection through learned nominal representations is vast, industrial image data comes with its own challenges \cite{mvtec}, for which recent works starting with \cite{bergman2020deep} have shown state-of-the-art detection performance using models pretrained on large external natural image datasets such as ImageNet \cite{imagenet} without any adaptation to the data at hand.
This has given rise to other industrial anomaly detection methods reliant on better reuse of pretrained features such as SPADE \cite{cohen2020transformerbased}, which utilizes memory banks comprising various feature hierarchies for finegrained, kNN-based \cite{eskin2002} anomaly segmentation and image-level anomaly detection.
Similarly, \cite{defard2020padim} recently proposed PaDiM, which utilizes a locally constrained bag-of-features approach \cite{brendel2018approximating}, estimating patch-level feature distribution moments (mean and covariance) for patch-level Mahalanobis distance measures \cite{mahalanobis1936generalized}. This approach is similar to \cite{rippel2020modeling} studied on full images.
To better account for the distribution shift between natural pretraining and industrial image data, subsequent adaptation can be done, e.g. via student-teacher knowledge distillation \cite{hinton2015distill} such as in  \cite{bergmann2020uninformed,salehi2020multiresolution} or normalizing flows \cite{dinh2016realnvp,kingma2018glow} trained on top of pretrained network features \cite{rudolph2021differnet}.

The specific components used in \PatchCore are most related to SPADE and PaDiM.
SPADE makes use of a memory-bank of nominal features extracted from a pretrained backbone network with separate approaches for image- and pixel-level anomaly detection.
\PatchCore similarly uses a memory bank, however with neighbourhood-aware patch-level features critical to achieve higher performance, as more nominal context is retained and a better fitting inductive bias is incorporated.
In addition, the memory bank is coreset-subsampled to ensure low inference cost at higher performance.
Coresets have seen longstanding usage in fundamental kNN and kMeans approaches \cite{harpeled2007knncoreset} or mixture models \cite{feldman2011mixturecoreset} by finding subsets that best approximate the structure of some available set and allow for approximate solution finding with notably reduced cost \cite{agarwal2004coresetgeometry,clarkson2010greedycoreset}. More recently, coreset-based methods have also found their way into Deep Learning approaches, e.g for network pruning \cite{mussay2020coresetpruning}, active learning \cite{sener2018coresetactive} and increasing effective data coverage of mini-batches for improved GAN training \cite{sinha2020smallgan} or representation learning \cite{roth2020revisiting}. The latter three have found success utilizing a greedy  coreset selection mechanism. As we aim to approximate memory bank feature space coverage, we similarly adapt a greedy coreset mechanism for \PatchCore.
Finally, our patch-level approach to both image-level anomaly detection and anomaly segmentation is related to PaDiM with the goal of encouraging higher anomaly detection sensitivity.
We make use of an efficient patch-feature memory bank equally accessible to all patches evaluated at test time, whereas PaDiM limits patch-level anomaly detection to Mahalanobis distance measures specific to each patch. In doing so, \PatchCore becomes less reliant on image alignment while also estimating anomalies using a much larger nominal context. Furthermore, unlike PaDiM, input images do not require the same shape during training and testing. Finally, \PatchCore makes use of locally aware patch-feature scores to  account for local spatial variance and to reduce bias towards ImageNet classes.

\section{Method}\label{sec:method}
\begin{figure*}
\centering
\includegraphics[width=0.93\textwidth]{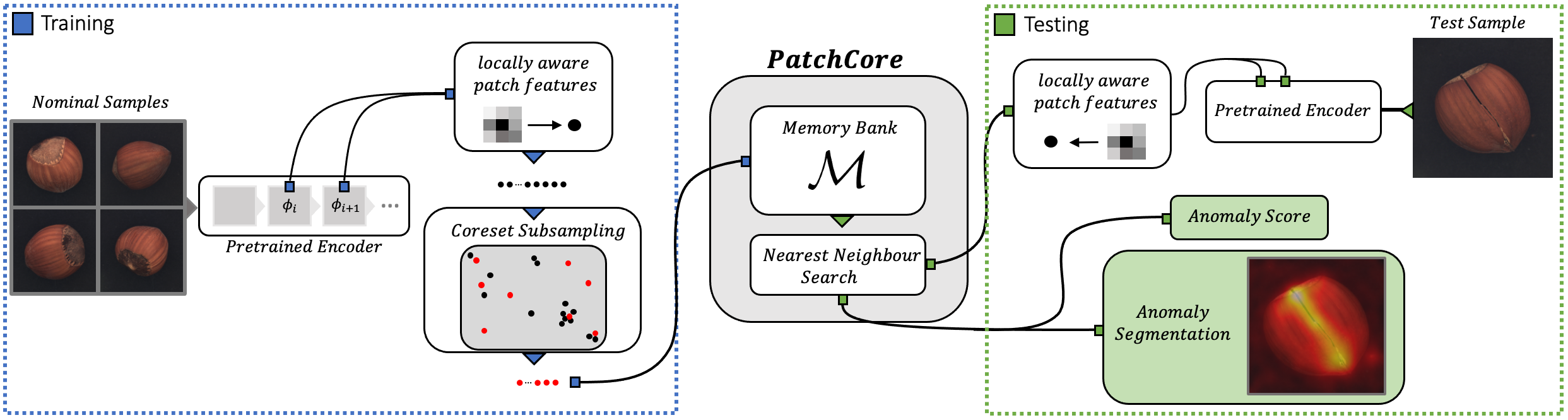}
\vspace{-3pt}
\caption{Overview of \textit{PatchCore}. Nominal samples are broken down into a memory bank of neighbourhood-aware patch-level features. For reduced redundancy and inference time, this memory bank is downsampled via greedy coreset subsampling. At test time, images are classified as anomalies if at least one patch is anomalous, and pixel-level anomaly segmentation is generated by scoring each patch-feature.}
\label{fig:patchcore_visualization}
\vspace{-3pt}
\end{figure*}

The \PatchCore method consists of several parts that we will describe in sequence: local patch features aggregated into a memory bank (\S\ref{subsec:prelim}), a coreset-reduction method to increase efficiency (\S\ref{subsec:memorybank}) and finally the full algorithm that arrives at detection and localization decisions (\S\ref{subsec:detection_and_localize}).

\subsection{Locally aware patch features}\label{subsec:prelim}
\noindent
We use $\mathcal{X}_N$ to denote the set of all nominal images ($\forall x \in \mathcal{X}_N: y_x = 0$) available at training time, with $y_{x} \in \{0, 1\}$ denoting if an image $x$ is nominal ($0$) or anomalous ($1$).  Accordingly, we define $\mathcal{X}_T$ to be the set of samples provided at test time, with $\forall x \in \mathcal{X}_T: y_x \in \{0, 1\}$.
Following~\cite{bergman2020deep}, \cite{cohen2020transformerbased} and \cite{defard2020padim}, \PatchCore uses a network $\phi$ pre-trained on ImageNet.
As the features at specific network hierarchies plays an important role, we use $\phi_{i, j} = \phi_j(x_i)$ to denote the features for image $x_i\in\mathcal{X}$ (with dataset $\mathcal{X}$) and hierarchy-level $j$ of the pretrained network $\phi$.
If not noted otherwise, in concordance with existing literature, $j$ indexes feature maps from ResNet-like \cite{He2015resnet} architectures, such as ResNet-50 or WideResnet-50 \cite{wideresnet}, with $j \in \{1, 2, 3, 4\}$ indicating the final output of respective spatial resolution blocks.

One choice for a feature representation would be the last level in the feature hierarchy of the network.
This is done in \cite{bergman2020deep} or \cite{cohen2020transformerbased} but introduces the following two problems.
Firstly, it loses more localized nominal information \cite{defard2020padim}. As the types of anomalies encountered at test time are not known \textit{a priori}, this becomes detrimental to the downstream anomaly detection performance.
Secondly, very deep and abstract features in ImageNet pretrained networks are biased towards the task of natural image classification, which has only little overlap with the cold-start industrial anomaly detection task and the evaluated data at hand.

We thus propose to use a memory bank $\mathcal{M}$ of patch-level features comprising \emph{intermediate} or \textit{mid-level} feature representations to make use of provided training context, avoiding features too generic or too heavily biased towards ImageNet classification. In the specific case of ResNet-like architectures, this would refer to e.g. $j\in[2, 3]$.
To formalize the patch representation we extend the previously introduced notation.
Assume the feature map $\phi_{i, j}\in\mathbb{R}^{c^*\times h^* \times w^*}$ to be a three-dimensional tensor of depth $c^*$, height $h^*$ and width $w^*$.
We then use $\phi_{i,j}(h, w) = \phi_j(x_i, h, w) \in\mathbb{R}^{c^*}$ to denote the $c^*$-dimensional feature slice at positions $h\in\{1,\ldots,h^*\}$ and $w\in\{1,\ldots,w^*\}$. Assuming the receptive field size of each $\phi_{i, j}$ to be larger than one, this effectively relates to image-patch feature representations.
Ideally, each patch-representation operates on a large enough receptive field size to account for meaningful anomalous context robust to local spatial variations. While this could be achieved by strided pooling and going further down the network hierarchy, the thereby created patch-features become more ImageNet-specific and thus less relevant for the anomaly detection task at hand, while training cost increases and effective feature map resolution drops.

This motivates a local neighbourhood aggregation when composing each patch-level feature representation to increase receptive field size and robustness to small spatial deviations without losing spatial resolution or usability of feature maps.
For that, we extend above notation for $\phi_{i,j}(h, w)$ to account for an uneven patchsizes $p$ (corresponding to the neighbourhood size considered), incorporating feature vectors from the neighbourhood
\begin{equation}\label{eq:neighbourhood}
\begin{split}
\mathcal{N}_p^{(h, w)} = \{(a, b) | &a \in [h - \lfloor p/2 \rfloor, ..., h + \lfloor p/2 \rfloor], \\&b \in  [w - \lfloor p/2 \rfloor, ..., w + \lfloor p/2 \rfloor]\},
\end{split}
\end{equation}
and locally aware features at position $(h, w)$ as
\begin{equation}
\phi_{i, j}\left(\mathcal{N}_p^{(h, w)}\right) = f_\text{agg}\left(\{\phi_{i, j}(a, b) | (a, b) \in \mathcal{N}_p^{(h, w)}\}\right),
\end{equation}
with $f_\text{agg}$ some aggregation function of feature vectors in the neighbourhood $\mathcal{N}_p^{(h, w)}$.
For \textit{PatchCore}, we use adaptive average pooling. This is similar to local smoothing over each individual feature map, and results in one single representation at $(h, w)$ of predefined dimensionality $d$, which is performed for all pairs $(h, w)$ with $h \in \{1, ..., h^*\}$ and $w\in\{1, ..., w^*\}$ and thus retains feature map resolution.
For a feature map tensor $\phi_{i, j}$, its locally aware patch-feature collection $\mathcal{P}_{s, p}(\phi_{i,j})$ is
\begin{equation}\label{eq:patchcollection}
\begin{split}
\mathcal{P}&_{s, p}(\phi_{i, j}) = \{ \phi_{i, j}(\mathcal{N}_p^{(h, w)}) | \\&h,w \text{ mod } s = 0, h < h^*, w < w^*, h, w \in \mathbb{N} \},
\end{split}
\end{equation}
with the optional use of a striding parameter $s$, which we set to $1$ except for ablation experiments done in \S\ref{subsubsec:subsampling}.
Empirically and similar to \cite{cohen2020transformerbased} and \cite{defard2020padim}, we found aggregation of multiple feature hierarchies to offer some benefit.
However, to retain the generality of used features as well as the spatial resolution, \PatchCore uses only two intermediate feature hierarchies $j$ and $j+1$.
This is achieved simply by computing $\mathcal{P}_{s, p}(\phi_{i, j+1})$ and aggregating each element with its corresponding patch feature at the lowest hierarchy level used (i.e., at the highest resolution), which we achieve by bilinearly rescaling $\mathcal{P}_{s, p}(\phi_{i, j+1})$ such that $|\mathcal{P}_{s, p}(\phi_{i, j+1})|$ and $|\mathcal{P}_{s, p}(\phi_{i, j})|$ match.

Finally, for all nominal training samples $x_i\in\mathcal{X}_N$, the \PatchCore memory bank $\mathcal{M}$ is then simply defined as
\begin{equation}
\mathcal{M} = \bigcup_{x_i\in\mathcal{X}_N} \mathcal{P}_{s, p}(\phi_j(x_i)).
\end{equation}

\subsection{Coreset-reduced patch-feature memory bank}\label{subsec:memorybank}
\begin{figure}
\centering
\begin{subfigure}[b]{0.28\textwidth}
\centering
\includegraphics[width=1\textwidth]{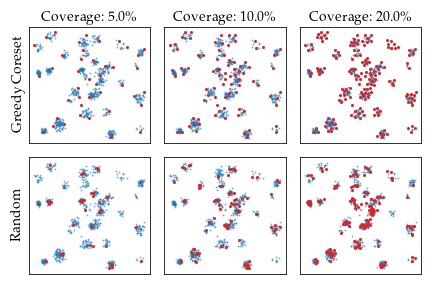}
\caption{}
\end{subfigure}
\begin{subfigure}[b]{0.19\textwidth}
\centering
\includegraphics[width=1\textwidth]{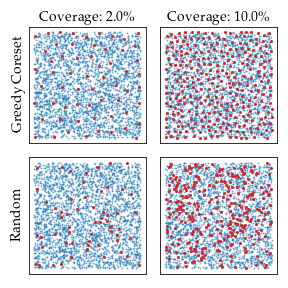}
\caption{}
\end{subfigure}
\vspace{-5pt}
\caption{Comparison: coreset (top) vs. random subsampling (bottom) (\red{red}) for 2D data (i\blue{blue}) sampled from (a) multimodal and (b) uniform distributions. Visually, coreset subsampling better approximates the spatial support, random subsampling misses clusters in the multi-modal case and is less uniform in (b).}
\label{fig:coreset_example}
\vspace{-3pt}
\end{figure}

For increasing sizes of $\mathcal{X}_N$, $\mathcal{M}$ becomes exceedingly large and with it both the inference time to evaluate novel test data and required storage. This issue has already been noted in SPADE \cite{cohen2020transformerbased} for anomaly segmentation, which makes use of both low- and high-level feature maps. Due to computational limitations, SPADE requires a preselection stage of feature maps for pixel-level anomaly detection based on a weaker image-level anomaly detection mechanism reliant on full-image, deep feature representations, i.e., global averaging of the last feature map.
This results in low-resolution, ImageNet-biased representations computed from full images which may negatively impact detection and localization performance.

These issues can be addressed by making $\mathcal{M}$ meaningfully searchable for larger image sizes and counts, allowing for patch-based comparison beneficial to both anomaly detection and segmentation.
This requires that the nominal feature coverage encoded in $\mathcal{M}$ is retained.
Unfortunately, random subsampling, especially by several magnitudes, will lose significant information available in $\mathcal{M}$ encoded in the coverage of nominal features (see also experiments done in \S\ref{subsubsec:subsampling}).
In this work we use a coreset subsampling mechanism to reduce $\mathcal{M}$, which we find reduces inference time while retaining performance.

Conceptually, coreset selection aims to find a subset $\mathcal{S}\subset\mathcal{A}$ such that problem solutions over $\mathcal{A}$ can be most closely and especially more quickly approximated by those computed over $\mathcal{S}$ \cite{agarwal2004coresetgeometry}.
Depending on the specific problem, the coreset of interest varies.
Because \PatchCore uses nearest neighbour computations (next Section), we use a \textit{minimax facility location} coreset selection, see e.g., \cite{sener2018coresetactive} and \cite{sinha2020smallgan}, to ensure approximately similar coverage of the $\mathcal{M}$-coreset $\mathcal{M}_C$ in patch-level feature space as compared to the original memory bank $\mathcal{M}$
\begin{equation}\label{eq:coreset}
\mathcal{M}_C^* = \argmin_{\mathcal{M}_C\subset\mathcal{M}} \max_{m\in\mathcal{M}}\min_{n\in\mathcal{M}_C} \left\Vert m - n \right\Vert_2.
\end{equation}
The exact computation of $\mathcal{M}_C^*$ is NP-Hard~\cite{wolsey2014integer}, we use the iterative greedy approximation suggested in \cite{sener2018coresetactive}.
To further reduce coreset selection time, we follow~\cite{sinha2020smallgan}, making use of the Johnson-Lindenstrauss theorem \cite{dasgupta2003johnsonlindenstrauss} to reduce dimensionalities of elements $m \in \mathcal{M}$ through random linear projections $\psi: \mathbb{R}^d \rightarrow \mathbb{R}^{d^*}$ with $d^* < d$.
The memory bank reduction is summarized in Algorithm \ref{alg:memory_bank}. For notation, we use \PatchCoreN{n} to denote the percentage $n$ to which the original memory bank has been subsampled to, e.g., \textit{PatchCore}$-1\%$ a 100x times reduction of $\mathcal{M}$. Figure \ref{fig:coreset_example} gives a visual impression of the spatial coverage of greedy coreset subsampling compared to random selection.

\begin{algorithm}
\SetAlgoLined
\KwInput{Pretrained $\phi$, hierarchies $j$, nominal data $\mathcal{X}_N$, stride $s$, patchsize $p$, coreset target $l$, random linear projection $\psi$.}
\KwOutput{Patch-level Memory bank $\mathcal{M}$.}
\textbf{Algorithm:}\\
 $\mathcal{M} \leftarrow \{\}$\\
 \For{$x_i\in\mathcal{X}_N$}{
 	$\mathcal{M} \leftarrow \mathcal{M} \cup \mathcal{P}_{s,p}(\phi_j(x_i))$
 }
 \tcc{\footnotesize{Apply greedy coreset selection.}}
$\mathcal{M}_C \leftarrow \{\}$\\
\For{$i\in[0, ..., l-1]$}{
	$m_i \leftarrow \underset{m\in\mathcal{M}-\mathcal{M}_C}{\argmax}\underset{n\in\mathcal{M}_C}{\min} \left\Vert \psi(m) - \psi(n) \right\Vert_2$\\
	$\mathcal{M}_C \leftarrow \mathcal{M}_C \cup \{m_i\}$\\
}
$\mathcal{M}\leftarrow\mathcal{M}_C$\\
\caption{\PatchCore memory bank.}
\label{alg:memory_bank}
\end{algorithm}

\subsection{Anomaly Detection with \textbf{\PatchCore}}\label{subsec:detection_and_localize}
With the nominal patch-feature memory bank $\mathcal{M}$, we estimate the image-level anomaly score $s\in\mathbb{R}$ for a test image $x^\text{test}$ by the maximum distance score $s^*$ between test patch-features in its patch collection $\mathcal{P}(x^\text{test}) = \mathcal{P}_{s, p}(\phi_j(x^\text{test}))$ to each respective nearest neighbour $m^*$ in $\mathcal{M}$:
\begin{equation}
\begin{split}
m^{\text{test}, *}, m^* &= \argmax_{m^\text{test} \in \mathcal{P}(x^\text{test})}\argmin_{m \in \mathcal{M}} \left\Vert m^\text{test} - m \right\Vert_2\\
s^* &= \left\Vert m^{\text{test},*} - m^* \right\Vert_2.
\end{split}
\end{equation}
To obtain $s$, we use scaling $w$ on $s^*$ to account for the behaviour of neighbour patches: If memory bank features closest to anomaly candidate $m^{\text{test}, *}$, $m^*$, are themselves far from neighbouring samples and thereby an already rare nominal occurrence, we increase the anomaly score
\begin{equation}\label{eq:image_level_anomaly_score}
s = \left(1-\frac{\exp{\left\Vert m^{\text{test}, *} - m^* \right\Vert_2}}{\sum_{m \in \mathcal{N}_b(m^*)} \exp{\left\Vert m^{\text{test},*} - m \right\Vert_2}}\right) \cdot s^*,
\end{equation}
with $\mathcal{N}_b(m^*)$ the $b$ nearest patch-features in $\mathcal{M}$ for test patch-feature $m^*$.
We found this re-weighting to be more robust than just the maximum patch distance.
Given $s$, segmentations follow directly.
The image-level anomaly score in Eq. \ref{eq:image_level_anomaly_score} (first line) requires the computation of the anomaly score for each patch through the $\argmax$-operation.
A segmentation map can be computed in the same step, similar to~\cite{defard2020padim}, by realigning computed patch anomaly scores based on their respective spatial location.
To match the original input resolution, (we may want to use intermediate network features), we upscale the result by bi-linear interpolation.
Additionally, we smoothed the result with a Gaussian of kernel width $\sigma=4$, but did not optimize this parameter.
%
%

\section{Experiments}\label{sec:experiments}


\subsection{Experimental Details}\label{subsec:implementation_details}
\begin{table*}[t]
\centering
\caption{Anomaly Detection Performance (AUROC) on MVTec AD \cite{mvtec}. PaDiM$^*$ denotes a result from \cite{defard2020padim} with problem-specific backbone selection. The total count of misclassifications was determined as the sum of false-positive and false-negative predictions given a F1-optimal threshold. We did not have individual anomaly scores for competing methods so could compute this number only for \textit{PatchCore}.}
\vspace{-3pt}
\resizebox{0.95\textwidth}{!}{
\begin{tabular}{|c||c|c|c|c|c|c||c|c|c|}
\toprule
\textbf{Method} & SPADE \cite{cohen2020transformerbased} & PatchSVDD \cite{yi2020patch}& DifferNet \cite{rudolph2021differnet}	& PaDiM \cite{defard2020padim}& Mah.AD \cite{rippel2020modeling}& PaDiM$^*$ \cite{defard2020padim}& PatchCore$-25\%$& PatchCore$-10\%$& PatchCore$-1\%$\\
\hline
\textbf{AUROC} $\uparrow$    &  85.5& 92.1& 94.9 & 95.3 & 95.8& 97.9& \textbf{99.1}& 99.0& 99.0\\
\textbf{Error}  $\downarrow$   &  14.5 & 7.9  & 5.1   & 4.7    & 4.2  & 2.1  & \textbf{0.9}& 1.0 & 1.0 \\
\textbf{Misclassifications} $\downarrow$  &  - &  - & - & - & - & - & \textbf{42} & 47 & 49\\
\bottomrule
\end{tabular}}
\label{tab:sota_mvtec_image_level}
\vspace{-3pt}
\end{table*}


\begin{table*}[t]
\caption{Anomaly Segmentation Performance (pixelwise AUROC) on MVTec AD \cite{mvtec}.}
\vspace{-3pt}
\centering
\resizebox{0.95\textwidth}{!}{
\begin{tabular}{|c||c|c|c|c|c|c||c|c|c|}
\toprule
\textbf{Method}    &  AE$_{SSIM}$	 \cite{mvtec} &  $\gamma$-VAE + grad. \cite{dehaene2020iterative}  & CAVGA-R$_w$ \cite{venkataramanan2019cavga}  & PatchSVDD \cite{yi2020patch}  & SPADE \cite{cohen2020transformerbased}  &  PaDiM \cite{defard2020padim}	&  PatchCore$-25\%$ & PatchCore$-10\%$  &  PatchCore$-1\%$ \\
\hline
\textbf{AUROC} $\uparrow$   &  87 & 88.8  &  89 &  95.7 & 96.0  & 97.5  &  \textbf{98.1} &   \textbf{98.1} &   98.0 \\
\textbf{Error}  $\downarrow$   &  13 & 11.2  & 11   & 4.3    & 4.0  & 2.5  & \textbf{1.9} & \textbf{1.9} & 2.0 \\
\bottomrule
\end{tabular}}
\label{tab:sota_mvtec_segmentation_auroc}
\vspace{-3pt}
\end{table*}


\begin{table*}[t]
\caption{Anomaly Detection Performance on MVTec AD \cite{mvtec}  as measured in PRO [$\%$] \cite{mvtec,cohen2020transformerbased}.}
\vspace{-4pt}
\centering
\resizebox{0.8\textwidth}{!}{
\begin{tabular}{|c||c|c|c|c||c|c|c|}
\toprule
\textbf{Method} & AE$_{SSIM}$	 \cite{mvtec} & Student \cite{bergmann2020uninformed} & SPADE \cite{cohen2020transformerbased} & PaDiM \cite{defard2020padim}  & PatchCore$-25\%$  & PatchCore$-10\%$  &  PatchCore$-1\%$ \\
\hline
\textbf{PRO} $\uparrow$ & 69.4 & 85.7 & 91.7 & 92.1 & 93.4 & \textbf{93.5} & 93.1 \\
\textbf{Error} $\downarrow$ & 30.6 & 14.3 & 8.3 & 7.9 & 6.6 & \textbf{6.5} & 6.9 \\
\bottomrule
\end{tabular}}
\label{tab:sota_mvtec_segmentation_pro}
\vspace{-4pt}
\end{table*}

\begin{table}[t]
\caption{PatchCore-$1\%$ with higher resolution/larger backbones/ensembles. The coreset subsampling allows for computationally expensive setups while still retaining fast inference.}
\centering
\resizebox{0.4\textwidth}{!}{
\begin{tabular}{|l||c|c|c|}
\toprule
\textbf{Metric}$\rightarrow$ & \textbf{AUROC} & \textbf{pwAUROC} & \textbf{PRO}\\
\midrule
\multicolumn{4}{l}{DenseN-201 \& RNext-101 \& WRN-101 (2+3), Imagesize 320}\\
\midrule
\textbf{Score} $\uparrow$ & \textbf{99.6} & 98.2 & 94.9 \\
\textbf{Error} $\downarrow$ & \textbf{0.4} & 1.8 & 5.6 \\
\midrule
\multicolumn{4}{l}{WRN-101 (2+3), Imagesize 280}\\
\midrule
\textbf{Score} $\uparrow$ & 99.4 & 98.2 & 94.4 \\
\textbf{Error} $\downarrow$ & 0.6 & 1.8 & 5.6 \\
\midrule
\multicolumn{4}{l}{WRN-101 (1+2+3), Imagesize 280}\\
\midrule
\textbf{Score} $\uparrow$ & 99.2 & \textbf{98.4} & \textbf{95.0} \\
\textbf{Error} $\downarrow$ & 0.8 & \textbf{1.6} & \textbf{5.0} \\
\bottomrule
\end{tabular}}
\label{tab:sota_optimized}
\end{table}


\begin{table}[h]
\centering
\caption{Mean inference time per image on MVTec AD. Scores are (image AUROC, pixel AUROC, PRO metric).}
\vspace{-3pt}
\resizebox{0.42\textwidth}{!}{
\begin{tabular}{|c|c|c|c|}
\toprule
\textbf{Method} & \textit{PatchCore}$-100\%$ & \textit{PatchCore}$-10\%$ & \textit{PatchCore}$-1\%$ \\ 
\hline
\textbf{Scores} & (99.1, 98.0, 93.3) & (99.0, 98.1, 93.5) & (99.0, 98.0, 93.1) \\
\textbf{Time} (s) & 0.6 & 0.22 & 0.17 \\
\midrule
\textbf{Method} & \textit{PatchCore}$-100\%$ + IVFPQ & SPADE & PaDiM \\
\hline
\textbf{Scores} & (98.0, 97.9, 93.0) & (85.3, 96.6, 91.5)& (95.4, 97.3, 91.8) \\
\textbf{Time} (s) & 0.2 & 0.66 & 0.19 \\
\bottomrule
\end{tabular}}
\label{tab:inference_times}
\vspace{-3pt}
\end{table}



\textbf{Datasets.}
To study industrial anomaly detection performance, the majority of our experiments are performed on the MVTec Anomaly Detection benchmark \cite{mvtec}.\\
MVTec AD contains 15 sub-datasets with a total of 5354 images, 1725 of which are in the test set.
Each sub-dataset is divided into nominal-only training data and test sets containing both nominal and anomalous samples for a specific product with various defect types as well as respective anomaly ground truth masks.
As in \cite{cohen2020transformerbased,defard2020padim,yi2020patch}, images are resized and center cropped to $256 \times 256$ and $224 \times 224$, respectively.
No data augmentation is applied, as this requires prior knowledge about class-retaining augmentations.\\
We also study industrial anomaly detection on more specialized tasks. For that, we leverage the \textit{Magnetic Tile Defects (MTD) \cite{magnetictiles}} dataset as used in \cite{rudolph2021differnet}, which contains 925 defect-free and 392 anomalous magnetic tile images with varied illumination levels and image sizes.
Same as in \cite{rudolph2021differnet}, $20\%$ of defect-free images are evaluated against at test time, with the rest used for cold-start training.\\
Finally, we also highlight potential applicability of \textit{PatchCore} to non-industrial image data, benchmarking cold-start anomaly localization on \textit{Mini Shanghai Tech Campus (mSTC)} as done in e.g. \cite{venkataramanan2019cavga} and \cite{defard2020padim}. \textit{mSTC} is a subsampled version of the original \textit{STC} dataset \cite{shanghaitechcampus}, only using every fifth training and test video frame. It contains pedestrian videos from 12 different scenes.
Training videos include normal pedestrian behaviour while test videos can contain different behaviours such as fighting or cycling.
For comparability of our cold-start experiments, we follow established \textit{mSTC} protocols \cite{venkataramanan2019cavga,defard2020padim}, not making use of any anomaly supervision and images resized to $256\times 256$. 
\noindent


\textbf{Evaluation Metrics.} Image-level anomaly detection performance is measured via the area under the receiver-operator curve (AUROC) using produced anomaly scores.
In accordance with prior work we compute on MVTec the class-average AUROC~\cite{akcay2018ganomaly,cohen2020transformerbased,defard2020padim}.
To measure segmentation performance, we use both pixel-wise AUROC and the PRO metric first, both following~\cite{bergmann2020uninformed}.
The PRO score takes into account the overlap and recovery of connected anomaly components to better account for varying anomaly sizes in MVTec AD, see~\cite{bergmann2020uninformed} for details.

\subsection{Anomaly Detection on MVTec AD}\label{subsec:mvtec}
\begin{figure}
\centering
\includegraphics[width=0.45\textwidth]{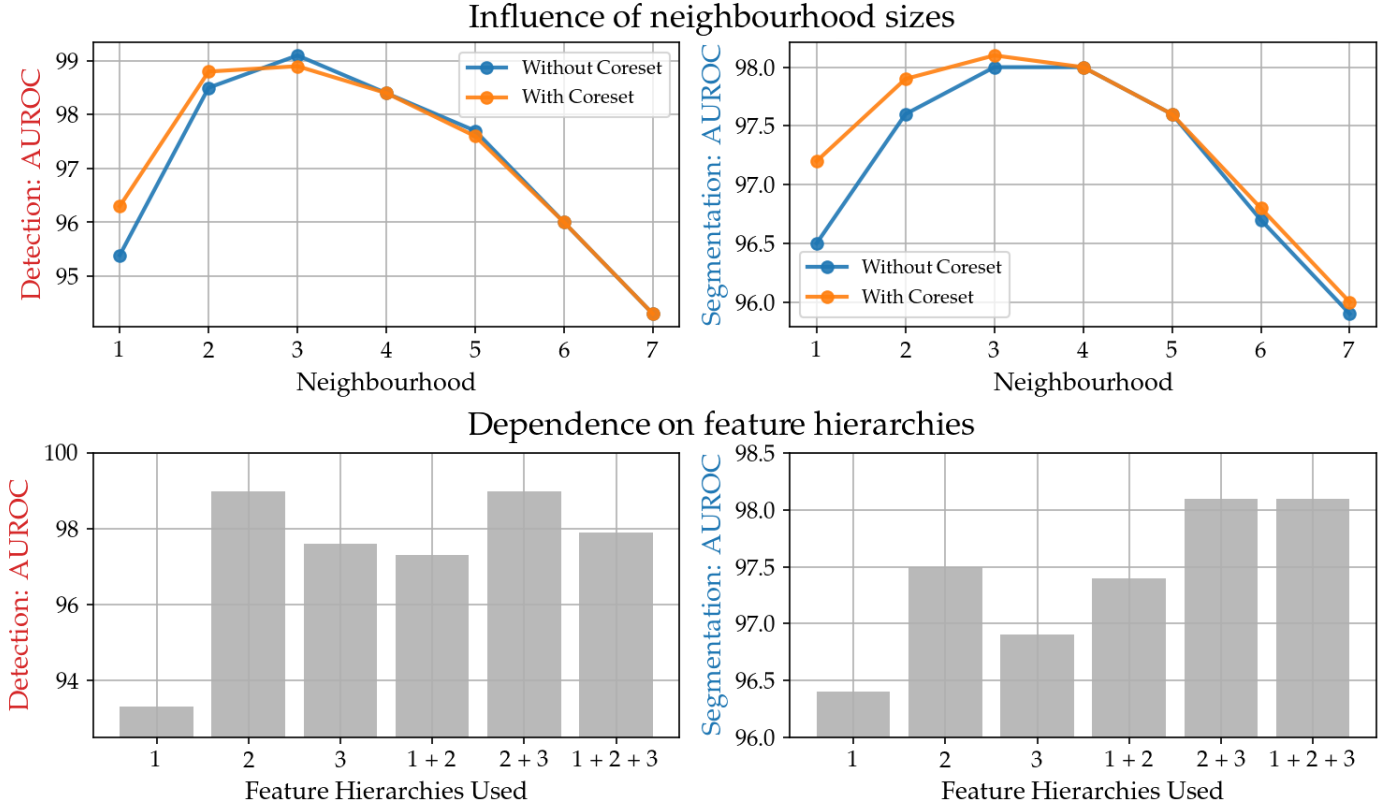}
\vspace{-3pt}
\caption{Local awareness and network feature depths vs. detection performance. PRO score results in the supplementary.}
\label{fig:method_ablation}
\vspace{-4pt}
\end{figure}

The results for image-level anomaly detection on MVTec are shown in Table \ref{tab:sota_mvtec_image_level}. For \PatchCore we report on various levels of memory bank subsampling ($25\%$, $10\%$ and $1\%$).
For all cases, \PatchCore achieves significantly higher mean image anomaly detection performance with consistently high performance on all sub-datasets (see supplementary \ref{appendix_section:complete_mvtec} for detailed comparison).
Please note, that a reduction from an error of 2.1\% (PaDiM) to 0.9\% for \PatchCoreN{25} means a reduction of the error by 57\%. In industrial inspection settings this is a relevant and significant reduction.
For MVTec at optimal F1 threshold, there are only 42 out of 1725 images classified incorrectly and a third of all classes are solved perfectly. 
In the supplementary material~\ref{appendix_section:complete_mvtec} we also show that both with F1-optimal working point and at full recall, classification errors are also lower when compared to both SPADE and PaDiM. With \textit{PatchCore}, less than 50 images remain misclassified.
In addition, \PatchCore achieves state-of-the-art anomaly segmentation, both measured by pixelwise AUROC (Table \ref{tab:sota_mvtec_segmentation_auroc}, $98.1$ versus $97.5$ for PaDiM) and PRO metric (Table \ref{tab:sota_mvtec_segmentation_pro}, $93.5$ versus $92.1$).
Sample segmentations in Figure \ref{fig:segmentation_examples} offer qualitative impressions of the accurate anomaly localization.

In addition, due to the effectiveness of our coreset memory subsampling, we can apply PatchCore$-1\%$ on images of higher resolution (e.g. $280/320$ instead of $224$) and ensemble systems while retaining inferences times less than PatchCore$-10\%$ on the default resolution. This allows us to further push image- and pixel-level anomaly detection as highlighted in Tab. \ref{tab:sota_optimized} (detailed results in supplementary), in parts more than halving the error again (e.g. $1\%\rightarrow 0.4\%$ for image-level AUROC). 

\subsection{Inference Time}
The other dimension we are interested in is inference time.
We report results in Table \ref{tab:inference_times} (implementation details in supp. \ref{appendix_section:implementation_details}) comparing to reimplementations of SPADE \cite{cohen2020transformerbased} and PaDiM \cite{defard2020padim} using WideResNet50 and operations on GPU where possible.
These inference times include the forward pass through the backbone.
As can be seen, inference time for joint image- and pixel-level anomaly detection of \PatchCoreN{100} (without subsampling) are lower than SPADE \cite{cohen2020transformerbased} but with higher performance.
With coreset subsampling, \textit{Patchcore} can be made even faster, with lower inference times than even PaDiM while retaining state-of-the-art image-level anomaly detection and segmentation performance.
Finally, we examine \PatchCoreN{100} with approximate nearest neighbour search (IVFPQ~\cite{faiss}) as an orthogonal way of reducing inference time (which can also be applied to SPADE, however which already performs notably worse than even \PatchCoreN{1}).
We find a performance drop, especially for image-level anomaly detection, while inference times are still higher than \PatchCoreN{1}.
Though even with performance reduction, approximate nearest neighbour search on \PatchCoreN{100} still outperforms other methods. A combination of coreset and approximate nearest neighbour would further reduce inference time, allowing scaling to much larger datasets.

\subsection{Ablations Study}\label{subsec:ablations}
We report on ablations for the locally aware patch-features and the coreset reduction method. Supplementary experiments show consistency across different backbones (\S\ref{appendix_subsec:pretraining}), scalability with increased image resolution (\S\ref{appendix_subsec:resolution}) and a qualitative analysis of remaining errors (\S\ref{appendix_subsec:misclassifications}).
\subsubsection{Locally aware patch-features and hierarchies}\label{subsubsec:method_ablation}
\begin{figure}
\centering
\includegraphics[width=0.42\textwidth]{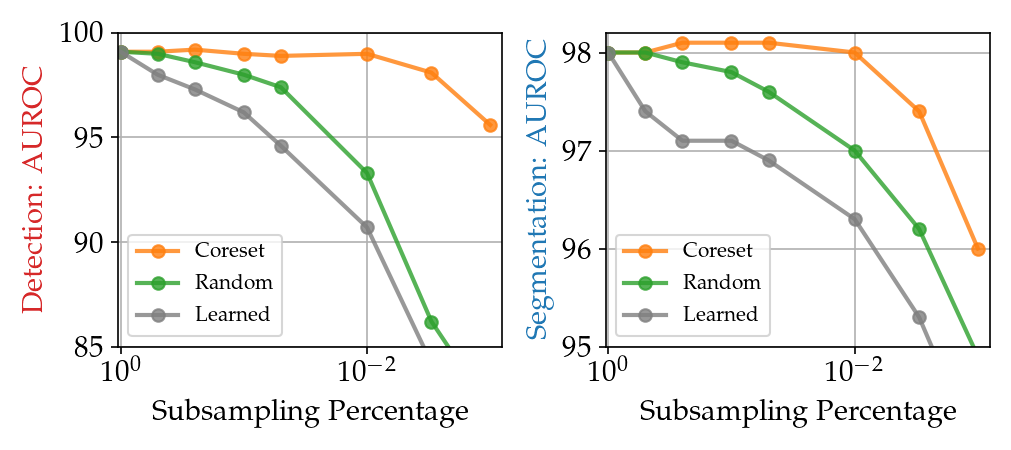}
\vspace{-5pt}
\caption{Performance retention for different subsamplers, results for PRO score in the supplementary.}
\label{fig:subsampling}
\vspace{-5pt}
\end{figure}



\begin{figure*}
  \centering
  \begin{subfigure}[b]{0.3\textwidth}
    \centering
    \includegraphics[width=1\textwidth]{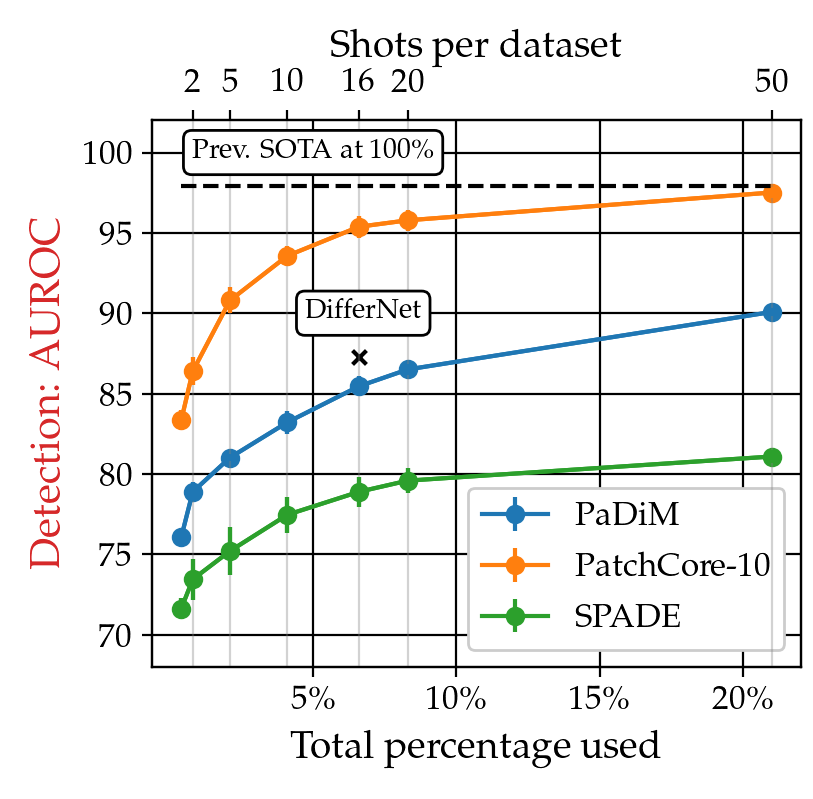}
  \end{subfigure}
  \begin{subfigure}[b]{0.3\textwidth}
    \centering
    \includegraphics[width=1\textwidth]{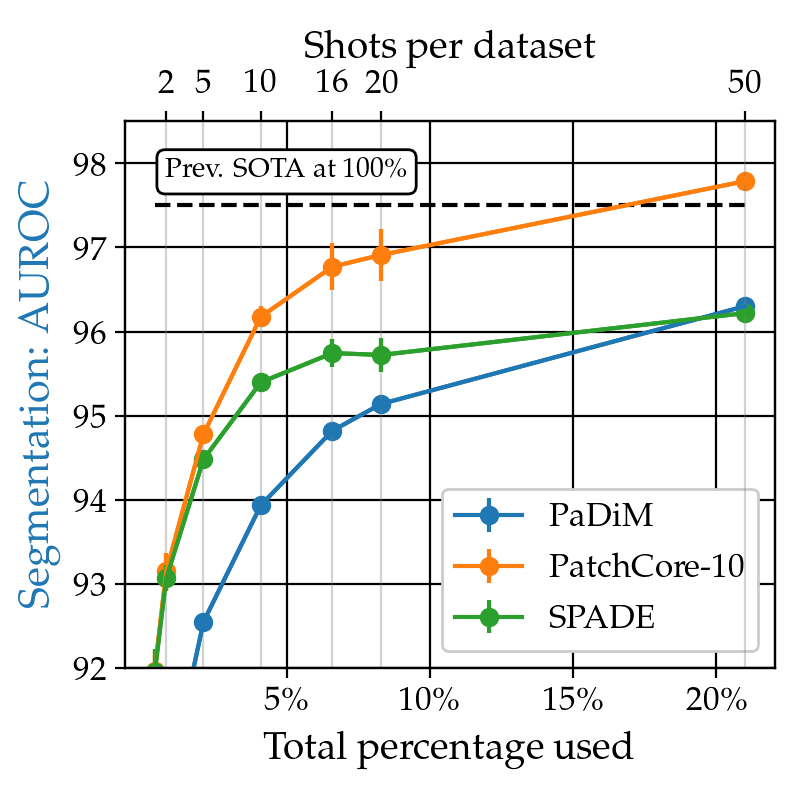}
  \end{subfigure}
  \begin{subfigure}[b]{0.3\textwidth}
    \centering
    \includegraphics[width=1\textwidth]{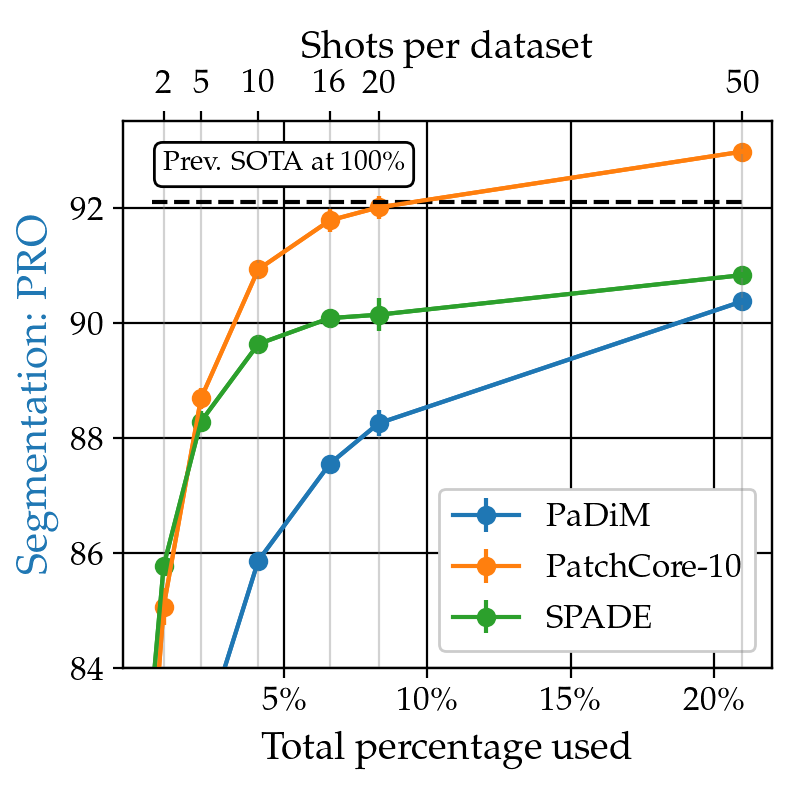}
  \end{subfigure}
\vspace{-5pt}
\caption{\textit{PatchCore} shows notably higher sample-efficiency than competitors, matching the previous state-of-the-art with a fraction of nominal training data. Note that PaDiM and SPADE where reimplemented with WideResNet50 for comparability.}
\label{fig:lowshot_performance}
\vspace{-7pt}
\end{figure*}

We investigate the importance of locally aware patch-features (\S\ref{subsec:detection_and_localize}) by evaluating changes in anomaly detection performance over different neighbourhood sizes in Eq. \ref{eq:neighbourhood}.
Results in the top half of Figure \ref{fig:method_ablation} show a clear optimum between locality and global context for patch-based anomaly predictions, thus motivating the neighbourhood size $p=3$.
More global context can also be achieved by moving down the network hierarchy (see e.g. \cite{cohen2020transformerbased,defard2020padim}), however at the cost of reduced resolution and heavier ImageNet class bias (\S\ref{subsec:prelim}). Indexing the first three WideResNet50-blocks with $1$ - $3$, Fig. \ref{fig:method_ablation} (\textit{bottom}) again highlights an optimum between highly localized predictions, more global context and ImageNet bias.
As can be seen, features from hierarchy level $2$ can already achieve state-of-the-art performance, but benefit from additional feature maps taken from subsequent hierarchy levels ($2 + 3$, which is chosen as the default setting).

\subsubsection{Importance of Coreset subsampling}\label{subsubsec:subsampling}
Figure \ref{fig:subsampling} compares different memory bank $\mathcal{M}$ subsampling methods: Greedy coreset selection, random subsampling and learning of a set of basis proxies corresponding to the subsampling target percentage $p_\text{target}$.
For the latter, we sample proxies $p_i\in\mathcal{P}\subset\mathbb{R}^d$ with $|\mathcal{P}| = p_\text{target} \cdot |\mathcal{M}|$,
which are then tasked to minimize a basis reconstruction objective
\begin{align}
	\mathcal{L}_\text{rec}(m_i) &= \left\Vert m_i - \sum_{p_k\in\mathcal{P}}\frac{e^{\left\Vert m_i - p_k\right\Vert_2}}{\sum_{p_j\in\mathcal{P}} e^{\left\Vert m_i - p_j\right\Vert}} p_k\right\Vert_2^2,
\end{align}
to find $N$ proxies that best describe the memory bank data $\mathcal{M}$.
In Figure \ref{fig:subsampling} we compare the three settings and find that coreset-based subsampling performs better than the other possible choices. The performance of no subsampling is comparable to a coreset-reduced memory bank that is two orders of magnitudes smaller in size. 
We also find subsampled memory banks to contain much less redundancy. We recorded the percentage of memory bank samples that are used at test time for non-subsampled and coreset-subsampled memory banks. While initially only less than $30\%$ of memory bank samples are used, coreset subsampling (to $1\%$) increases this factor to nearly $95\%$.
For certain subsampling intervals (between around $50\%$ and $10\%$), we even find joint performance over anomaly detection and localization to partly increase as compared to non-subsampled \PatchCore.
Finally, reducing the memory bank size $\mathcal{M}$ by means of increased striding (see Eq. \ref{eq:patchcollection}) shows worse performance due to the decrease in resolution context, with stride $s=2$ giving an image anomaly detection AUROC of $97.6\%$, and stride $s=3$ an AUROC of $96.8\%$.
\subsection{Low-shot Anomaly Detection}\label{subsec:lowshot}
Having access to limited nominal data is a relevant setting for real-world inspection. Therefore in addition to reporting results on the full MVTec AD, we also study the performance with fewer training examples. 
We vary the amount of training samples from 1 (corresponding to $0.4\%$ of the total nominal training data) to 50 ($21\%$), and compare to reimplementations of SPADE \cite{cohen2020transformerbased} and PaDiM \cite{defard2020padim} using the same backbone (WideResNet50).
Results are summarized in Figure \ref{fig:lowshot_performance}, with detailed results available in Supp. \S\ref{appendix_subsec:detailed_lowshot}.
As shown, using only one fifth of nominal training data, \PatchCore can still match previous state-of-the-art performance. In addition, comparing to the 16-shot experiments performed in \cite{rudolph2021differnet}, we find \PatchCore to outperform their approach which adapts a normalizing flows model on top of already pretrained features.
Compared to image-level memory approaches in \cite{cohen2020transformerbased}, we find matching localization and detection performance with only $5$/$1$ nominal shots.


\subsection{Evaluation on other benchmarks}\label{subsec:other_benchmarks}
\begin{table}[t]
\centering
\caption{Anomaly Segmentation on mSTC \cite{venkataramanan2019cavga,shanghaitechcampus} and anomaly detection on MTD \cite{magnetictiles} compared to results reported in \cite{rudolph2021differnet}.}
\vspace{-4pt}
\resizebox{0.44\textwidth}{!}{
\begin{tabular}{|c||c|c|c||c|}
\toprule
\textbf{mSTC}&CAVGA-R$_u$ \cite{venkataramanan2019cavga}& SPADE \cite{cohen2020transformerbased} & PaDiM \cite{defard2020padim} &  \textit{PatchCore}$-10$\\
\hline
\textit{Pixelwise AUROC} [$\%$] & 85 & 89.9 & 91.2 & \textbf{91.8} \\
\midrule
\textbf{MTD}&GANomaly \cite{akcay2018ganomaly}& 1-NN \cite{nazare2018nn1}&DifferNet \cite{rudolph2021differnet} & \textit{PatchCore}$-10$\\
\hline
\textit{AUROC} [$\%$] & 76.6 & 80.0 & 97.7 & \textbf{97.9} \\
\bottomrule
\end{tabular}}
\label{tab:sota_stc_mtd_results}
\vspace{-8pt}
\end{table}
We benchmark \PatchCore on two additional anomaly detection performance benchmarks: The ShanghaiTech Campus dataset (STC)~\cite{shanghaitechcampus} and the Magnetic Tile Defects dataset (MTD)~\cite{magnetictiles}.
Evaluation for STC as described in \S\ref{subsec:implementation_details} follows \cite{venkataramanan2019cavga}, \cite{defard2020padim} and \cite{cohen2020transformerbased}. We report unsupervised anomaly localization performance on a subsampled version of the STC video data (mSTC), with images resized to $256\times 256$ \cite{defard2020padim}.
As the detection context is much closer to natural image data available in ImageNet, we make use of deeper network feature maps at hierarchy levels 3 and 4, but otherwise do not perform any hyperparameter tuning for \textit{PatchCore}. The results in Table \ref{tab:sota_stc_mtd_results} (\textit{top}) show state-of-the-art anomaly localization performance which suggests good transferability of \PatchCore to such domains.
Finally, we examine MTD, containing magnetic tile defect images of varying sizes on which spatially rigid approaches like PaDiM cannot be applied directly. Here, nominal data already exhibits high variability similar to those encountered in anomalous samples~\cite{rudolph2021differnet}. We follow the protocol proposed in~\cite{rudolph2021differnet} to measure image-level anomaly detection performance and find performance to match (and even slightly outperform) that of~\cite{rudolph2021differnet} (Table \ref{tab:sota_stc_mtd_results}, \textit{bottom}).

\section{Conclusion}
This paper introduced the \PatchCore algorithm for cold-start anomaly detection, in which knowledge of only nominal examples has to be leveraged to detect and segment anomalous data at test-time.
\PatchCore strikes a balance between retaining a maximum amount of nominal context at test-time through the usage of memory banks comprising locally aware, nominal patch-level feature representations extracted from ImageNet pretrained networks, and minimal runtime through coreset subsampling. 
The result is a state-of-the-art cold-start image anomaly detection and localization system with low computational cost on industrial anomaly detection benchmarks. On MVTec, we achieve an image anomaly detection AUROC over $99\%$ with highest sample efficiency in relevant small training set regimes. 

\textbf{Broader Impact.} As automated industrial anomaly detection is one of the most successful applications of Computer Vision, the improvements gained through \PatchCore can be of notable interest for practitioners in this domain. As our work focuses specifically on industrial anomaly detection, negative societal impact is limited. And while the fundamental approach can potentially we leveraged for detection systems in more controversial domains, we don't believe that our improvements are significant enough to change societal application of such systems.

\textbf{Limitations.} While \PatchCore shows high effectiveness for industrial anomaly detection without the need to specifically adapt to the problem domain at hand, applicability is generally limited by the transferability of the pretrained features leveraged. This can be addressed by merging the effectiveness of \PatchCore with adaptation of the utilized features. We leave this interesting extension to future work.

\section*{Acknowledgements}
We thank Yasser Jadidi and Alex Smola for setup support of our compute infrastructure. K.R. thanks the International Max Planck Research School for Intelligent Systems (IMPRS-IS) and the European Laboratory for Learning
and Intelligent Systems (ELLIS) PhD program for support.



{\small
\bibliographystyle{ieee_fullname}
\bibliography{references}
}

\clearpage
\onecolumn

\begin{centering}
\section*{\Large{Supplementary:\\Towards Total Recall in Industrial Anomaly Detection}}
\end{centering}
\vspace{12pt}

\setcounter{equation}{0}
\setcounter{figure}{0}
\setcounter{table}{0}
\setcounter{page}{1}
\makeatletter
\renewcommand{\theequation}{S\arabic{equation}}
\renewcommand{\thefigure}{S\arabic{figure}}
\renewcommand{\thetable}{S\arabic{table}}

\appendix
\section{Implementation Details}\label{appendix_section:implementation_details}
We implemented our models in Python 3.7 \cite{python} and PyTorch \cite{pytorch}.
Experiments are run on Nvidia Tesla V4 GPUs. 
We used  torchvision ImageNet-pretrained models from torchvision and the PyTorch Image Models repository \cite{rw2019timm}. By default, following \cite{cohen2020transformerbased} and \cite{defard2020padim}, \PatchCore uses a WideResNet50-backbone \cite{wideresnet} for direct comparability. Patch-level features are taken from feature map aggregation of the final outputs in blocks $2$ and $3$. For all nearest neighbour retrieval and distance computations, we use \texttt{faiss} \cite{faiss}.

\section{Full MVTec AD comparison}\label{appendix_section:complete_mvtec}
This section contains a more detailed comparison on MVTec AD.
We include more models and a more finegrained performance comparison on all MVTec AD sub-datasets where available.
In the main part of the paper this has been referenced in \S\ref{subsec:mvtec}.
The corresponding result tables are \ref{tab:appendix_mvtec_sota_img_auroc}, \ref{tab:appendix_mvtec_sota_seg_auroc} and \ref{tab:appendix_mvtec_sota_seg_pro}.
We observe that \PatchCoreN{25} solves six of the 15 MVTec datasets and achieves highest AUROC performance on most datasets and in average. 

Figure \ref{fig:curves} show Precision-Recall and ROC curves for \PatchCore variants as well as reimplemented, comparable methods SPADE \cite{cohen2020transformerbased} and PaDiM \cite{defard2020padim} using a WideResNet50 backbone.
We also plot classification error both at 100\% recall and under a F1-optimal threshold to give a comparable working point. As can be seen, \PatchCore achieves consistently low classification errors with defined working points as well, with near-optimal Precision-Recall and ROC curves across datasets, in contrast to SPADE and PaDiM.

Finally, Table \ref{tab:appendix_mvtec_detailed_extended} showcases the detailed performance on all MVTec AD subdatasets for larger imagesizes ($280 \times 280$) and a WideResNet-101 backbone for further performance boosts using PatchCore$-1\%$, which allows for efficient anomaly detection at inference time even with larger images.

\section{Additional Ablations \& Details}
\subsection{Detailed Low-Shot experiments}\label{appendix_subsec:detailed_lowshot}
This section offers detailed numerical values to the low-shot method study provided in the main part of this work (\S\ref{subsec:lowshot}).
The results are included in Table~\ref{tab:lowshot_mvtec} and we find consistently higher numbers for detection and anomaly localization metrics.

\subsection{Dependency on pretrained networks}\label{appendix_subsec:pretraining}
We tested \PatchCore with different backbones, the results are shown in \ref{tab:backbones_mvtec}. We find that results are mostly stable over the choice of different backbones. The choice of WideResNet50 was made to be comparable with SPADE and PaDiM.

\subsection{Influence of image resolution}\label{appendix_subsec:resolution}
Next we study the influence of image size on performance. 
In the main paper we have used $224\times 224$ to be comparable with prior work.
In Figure~\ref{fig:resolution} we vary the image size from $288\times 288$, $360\times 360$ to $448\times 448$ and the neighborhood sizes (P) within 3, 5, 7, and 9.
We observe slightly increased detection performance and the performance saturates for~\PatchCore. 
For anomaly segmentation we observe a consistent increase, so if good localization is of importance, this is an ingredient to validate over. 
%

\subsection{Remaining Misclassifications}\label{appendix_subsec:misclassifications}
\begin{figure*}
\centering
\includegraphics[width=1\textwidth]{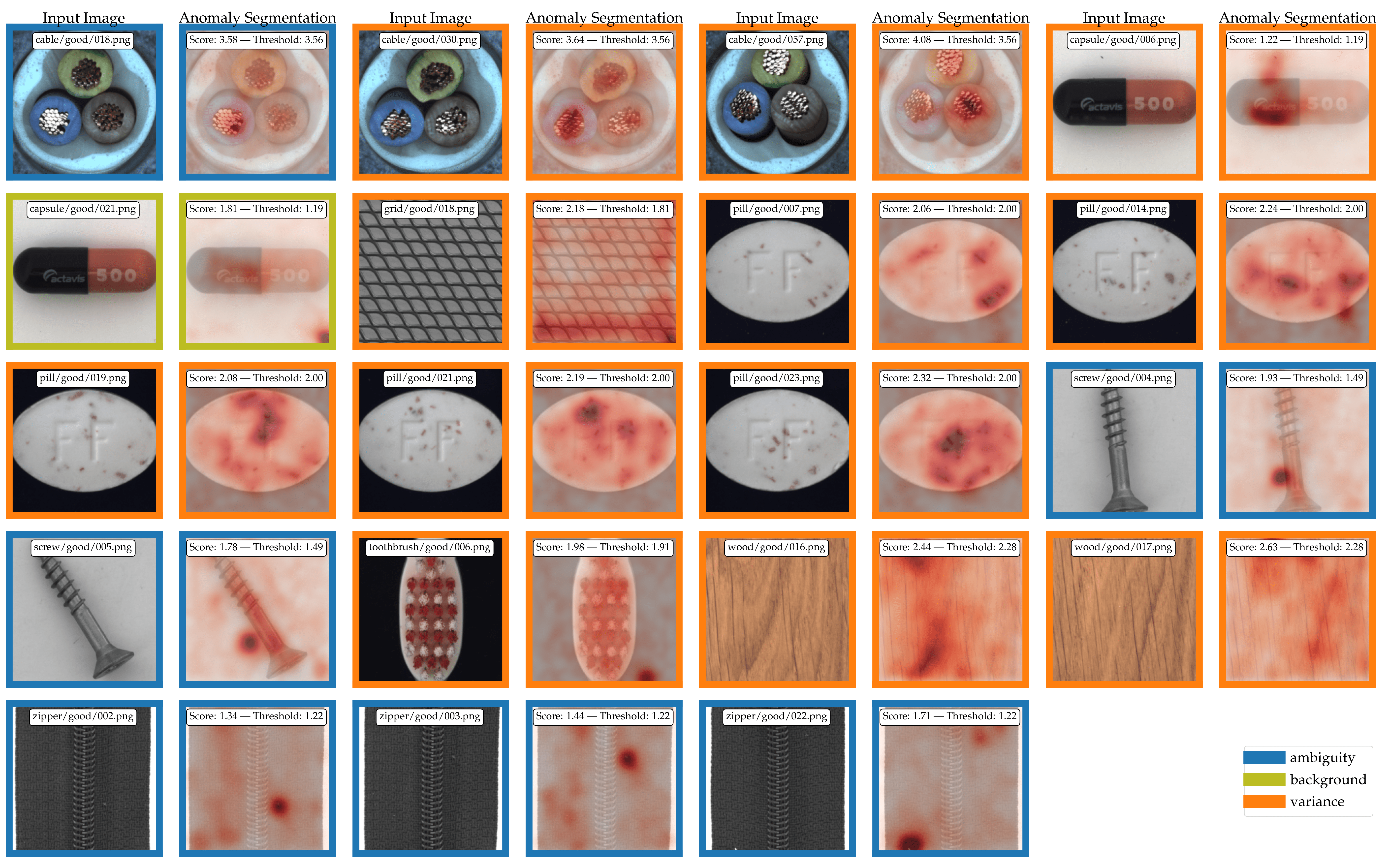}
\caption{Visualization of remaining false positive classifications (under F1-optimal thresholding). Colors denote different error sources. \orange{Orange} denotes high degrees of nominal variance mistaken for anomalies, \blue{blue} denotes misclassifications due to anomalies in the labelling context and \olive{olive} denotes variance in the background mistaken for anomalous content.}
\label{fig:remaining_errors_pos}
\end{figure*}

\begin{figure*}
\centering
\includegraphics[width=1\textwidth]{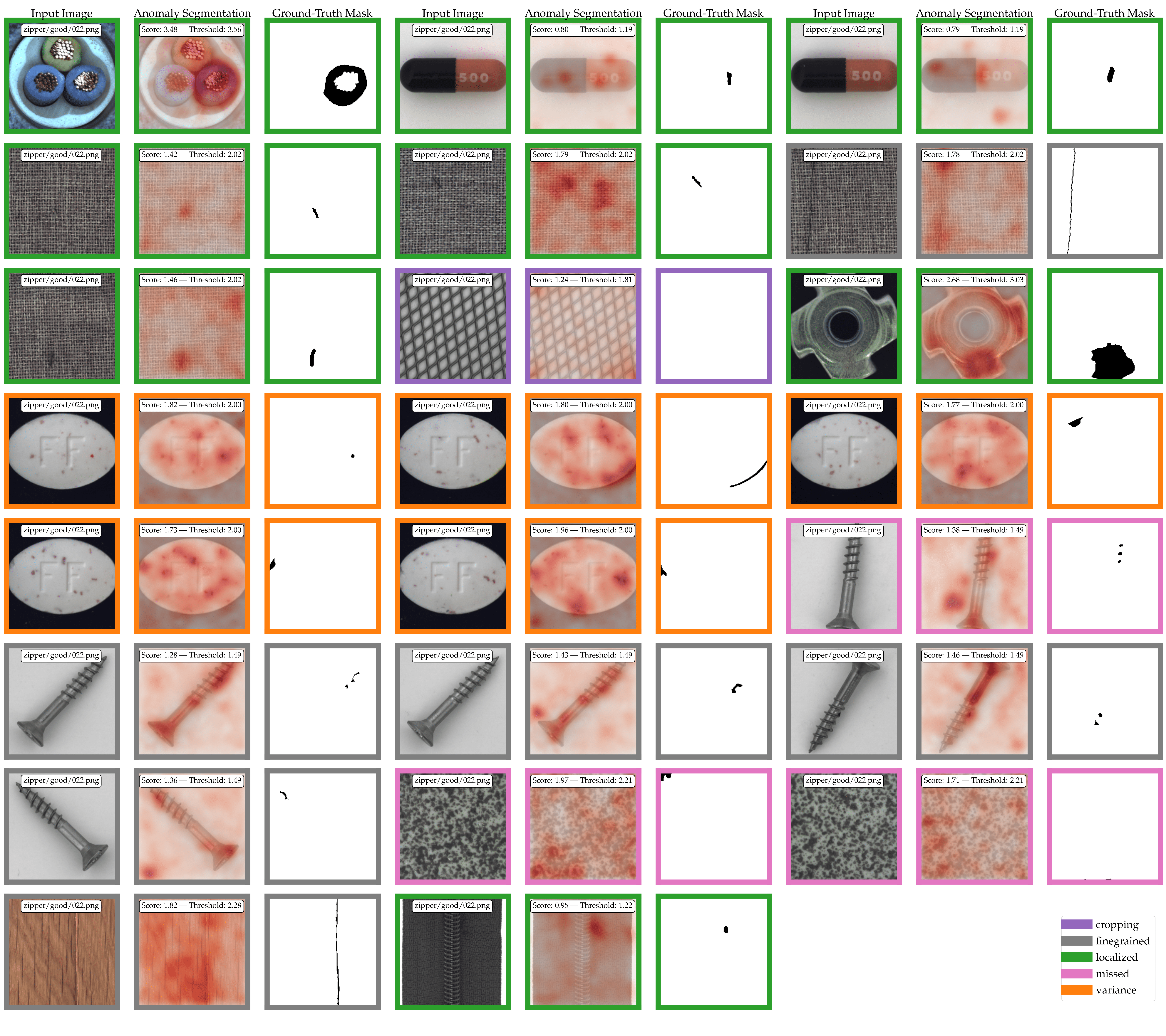}
\caption{Visualization of remaining false-negative classifications (under F1-optimal thresholding). Colors denote different error sources. \orange{Orange} denotes high degrees of nominal variance mistaken for anomalies, \green{green} denotes actually localized anomalies, but too little weight placed on these anomalies, \pink{pink} stands for anomalies that were not recovered, \purple{purple} denotes anomalies missed due to cropping-based image-processing (one anomaly in total), and \gray{gray} stands for finegrained anomalies that could be recovered when operating on higher image resolutions.}
\label{fig:remaining_errors_neg}
\end{figure*}
The high image-level anomaly detection performance allows us to look into all remaining misclassifications in detail. 
We compute the working point (threshold above which scores are considered anomalous) using the F1-optimal point.
With this threshold a total of 19 false-positive and 23 false-negative errors remain, all of which are visualized in Figures \ref{fig:remaining_errors_pos} and \ref{fig:remaining_errors_neg}.
Each segmentation map was normalized to the threshold value, so in some cases background scores are pronounced disproportionally.

Looking at Figure \ref{fig:remaining_errors_pos}, we find that the majority of false-positive errors either stem from a) (in \blue{blue}) ambiguity in labelling , i.e., image changes that could also be potentially labelled as anomalous, and b) (in \orange{orange}) very high nominal variance, resembling potential anomalies .
While the former can hardly be addressed by proposed methods, the latter could be addressed by offering some form of adaptation to the nominal data.
However, as \PatchCore outperforms adaptive methods, such adaptation would show most promise operating alongside pretraining-based methods such as \PatchCore.

To understand false-negative errors made, we include in Figure \ref{fig:remaining_errors_neg} the generated segmentation maps and ground-truth masks.
As can be seen, a large part of anomalies are localized well, however with insufficient weight placed on the anomalous regions, and could potentially be addressed by some means of postprocessing.
Other misclassifications are caused mostly by either high degrees of nominal variance that gets mistaken for anomalous context, and finegrained anomalies that could be captured when moving to higher image resolutions.
The amount of completely missed anomalies is small in comparison, and in one case caused by image preprocessing cropping out the actual anomalous region.

\subsection{Local Awareness and Subsampling}
\begin{table*}[t]
\centering
\caption{Anomaly Detection Performance (AUROC) on MVTec AD \cite{mvtec}. PaDiM$^*$ denotes a result from \cite{defard2020padim} with a backbone specifically selected for the task of image-level anomaly detection, which we could not reproduce.}
\resizebox{\textwidth}{!}{
\begin{tabular}{|l||c||c|c|c|c|c|c|c|c|c|c|c|c|c|c|c|}
\toprule
$\downarrow$ Method \textbackslash Dataset $\rightarrow$& \textbf{Avg} & Bottle & Cable & Capsule & Carpet & Grid & Hazeln. & Leather & Metal Nut & Pill & Screw & Tile & Toothb. & Trans. & Wood & Zipper\\
\midrule
GeoTrans \cite{izhak2018geometric}						& 67.2 & 74.4 & 78.3 & 67.0 & 43.7 & 61.9 & 35.9 & 84.1 & 81.3 & 63.0 & 50.0 & 41.7 & 97.2 & 86.9 & 61.1 &  82.0\\
GANomaly \cite{akcay2018ganomaly}					& 76.2 & 89.2 & 75.7 & 73.2 & 69.9 & 70.8 & 78.5 & 84.2 & 70.0 & 74.3 & 74.6 & 79.4 & 65.3 & 79.2 & 83.4 & 74.5\\
DSEBM \cite{zhai2016dsebm}								& 70.9 & 81.8 & 68.5 & 59.4 & 41.3 & 71.7 & 76.2 & 41.6 & 67.9 & 80.6 & 99.9 & 69.0 & 78.1 & 74.1 & 95.2 & 58.4\\
OCSVM \cite{andrews2016ocsvm}							& 71.9 & 99.0 & 80.3 & 54.4 & 62.7 & 41.0 & 91.1 & 88.0 & 61.1 & 72.9 & 74.7 & 87.6 & 61.9 & 56.7 & 95.3 & 51.7\\
ITAE \cite{huang2019inversetransform} 					& 83.9 & 94.1 & 83.2 & 68.1 & 70.6 & 88.3 & 85.5 & 86.2 & 66.7 & 78.6 & \textbf{100} & 73.5 & \textbf{100} & 84.3 & 92.3 & 87.6\\
SPADE \cite{cohen2020transformerbased}				& 85.5 & - & - & - & - & - & - & - & - & - & - & - & - & - & - & - \\
CAVGA-R$_w$ \cite{venkataramanan2019cavga} 	& 90 & 96 & 92 & 93 & 88 & 84 & 97 & 89 & 82 & 86 & 81 & 97 & 89 & 99 & 79 & 96\\
PatchSVDD \cite{yi2020patch}								& 92.1 & 98.6 & 90.3 & 76.7 & 92.9 & 94.6 & 92.0 & 90.9 & 94.0 & 86.1 & 81.3 & 97.8 & \textbf{100} & 91.5 & 96.5 & 97.9\\
DifferNet \cite{rudolph2021differnet}						& 94.9 & 99.0 & 95.9 & 86.9 & 92.9 & 84.0 & 99.3 & 97.1 & 96.1 & 88.8 & 96.3 & 99.4 & 98.6 & 91.1 & \textbf{99.8} & 95.1 \\
PaDiM \cite{defard2020padim}								& 95.3 & - & - & - & - & - & - & - & - & - & - & - & - & - & - & - \\
MahalanobisAD \cite{rippel2020modeling} 				& 95.8 & \textbf{100} & 95.0 & 95.1 & \textbf{100} & 89.7 & 99.1 & \textbf{100} & 94.7 & 88.7 & 85.2 & \textbf{99.8} & 96.9 & 95.5 & 99.6 & 97.9\\
PaDiM$^*$ \cite{defard2020padim} 						& 97.9 & - & - & - & - & - & - & - & - & - & - & - & - & - & - & - \\
\midrule
PatchCore$-25$ & \blue{\textbf{99.1}} & \textbf{100} & \textbf{99.5} & \textbf{98.1} & \textbf{98.7} & 98.2 & \textbf{100} & \textbf{100} & \textbf{100} & 96.6 & 98.1 & 98.7 & \textbf{100} & \textbf{100} & 99.2 & 99.4 \\
PatchCore$-10$ & 99.0                          & 100 & 99.4 & 97.8 & 98.7 & 97.9 & \textbf{100} & \textbf{100} & \textbf{100} & 96.0 & 97.0 & 98.9 & 99.7 & \textbf{100} & 99.0 & \textbf{99.5} \\
PatchCore$-1$   & 99.0 					   & 100 & 99.3 & 98.0 & 98.0 & \textbf{98.6} & \textbf{100} & \textbf{100} & 99.7 & \textbf{97.0} & 96.4 & 99.4 & \textbf{100} & 99.9 & 99.2 & 99.2 \\

\bottomrule
\end{tabular}}
\label{tab:appendix_mvtec_sota_img_auroc}
\end{table*}

\begin{table*}[t]
\centering
\caption{Anomaly Segmentation Performance on MVTec \cite{mvtec}, as measured in pixelwise AUROC.}
\resizebox{\textwidth}{!}{
\begin{tabular}{|l||c||c|c|c|c|c|c|c|c|c|c|c|c|c|c|c|}
\toprule
$\downarrow$ Method \textbackslash Dataset $\rightarrow$&  \textbf{Avg} & Bottle & Cable & Capsule & Carpet & Grid & Hazeln. & Leather & Metal Nut & Pill & Screw & Tile & Toothb. & Trans. & Wood & Zipper\\
\midrule
vis. expl. VAE \cite{liu2020vevae}  								& 86 & 87 & 90 & 74 & 78 & 73 & 98 & 95 & 94 & 83 & 97 & 80 & 94 & 93 & 77 & 78 \\
AE$_{SSIM}$	 \cite{mvtec}											& 87 & 93 & 82 & 94 & 87 & 94 & 97 & 78 & 89 & 91 & 96 & 59 & 92 & 90 & 73 & 88\\
$\gamma$-VAE + grad. \cite{dehaene2020iterative} 	& 88.8 & 93.1 & 88.0 & 91.7 & 72.7 & 97.9 & 98.8 & 89.7 & 91.4 & 93.5 & 97.2 & 58.1 & 98.3 & 93.1 & 80.9 & 87.1 \\
CAVGA-R$_w$ \cite{venkataramanan2019cavga} 		& 89 & - & - & - & - & - & - & - & - & - & - & - & - & - & - & -\\
PatchSVDD \cite{yi2020patch}									& 95.7 & 98.1 & 96.8 & 95.8 & 92.6 & 96.2 & 97.5 & 97.4 & 98.0 & 95.1 & 95.7 & 91.4 & 98.1 & 97.0 & 90.8 & 95.1 \\
SPADE \cite{cohen2020transformerbased} 				& 96.0 & 98.4 & 97.2 & \textbf{99.0} & 97.5 & 93.7 & \textbf{99.1} & 97.6 & 98.1 & 96.5 & 98.9 & 87.4 & 97.9 & 94.1 & 88.5 & 96.5 \\
PaDiM \cite{defard2020padim}									& 97.5 & 98.3 & 96.7 & 98.5 & \textbf{99.1} & 97.3 & 98.2 & 99.2 & 97.2 & 95.7 & 98.5 & 94.1 & \textbf{98.8} & \textbf{98.5} & 94.9 & 98.5  \\
\midrule
PatchCore$-25$ & \blue{\textbf{98.1}} & \textbf{98.6} & 98.4 & 98.8 & 99.0 & \textbf{98.7} & 98.7 &\textbf{ 99.3 }& \textbf{98.4} & 97.4 & \textbf{99.4} & 95.6 & 98.7 & 96.3 & 95.0 & 98.8 \\
PatchCore$-10$ & \blue{\textbf{98.1}} & \textbf{98.6} & \textbf{98.5} & 98.9 & \textbf{99.1} & \textbf{98.7} & 98.7 & \textbf{99.3} & \textbf{98.4} & \textbf{97.6} & \textbf{99.4} & 95.9 & 98.7 & 96.4 & \textbf{95.1} & \textbf{98.9} \\
PatchCore$-1$   & 98.0 & 98.5 & 98.2 & 98.8 & 98.9 & 98.6 & 98.6 & \textbf{99.3} & \textbf{98.4} & 97.1 & 99.2 & \textbf{96.1} & 98.5 & 94.9 & \textbf{95.1} & 98.8 \\
\bottomrule
\end{tabular}}
\label{tab:appendix_mvtec_sota_seg_auroc}
\end{table*}

\begin{table*}[t]
\centering
\caption{Anomaly Segmentation Performance on MVTec \cite{mvtec}, as measured in PRO [$\%$] \cite{mvtec,cohen2020transformerbased}.}
\resizebox{\textwidth}{!}{
\begin{tabular}{|l||c||c|c|c|c|c|c|c|c|c|c|c|c|c|c|c|}
\toprule
$\downarrow$ Method \textbackslash Dataset $\rightarrow$&  \textbf{Avg} & Bottle & Cable & Capsule & Carpet & Grid & Hazeln. & Leather & Metal Nut & Pill & Screw & Tile & Toothb. & Trans. & Wood & Zipper\\
\midrule
AE$_{SSIM}$	 \cite{mvtec}											& 69.4 & 83.4 & 47.8 & 86.0 & 64.7 & 84.9 & 91.6 & 56.1 & 60.3 & 83.0 & 88.7 & 17.5 & 78.4 & 72.5 & 60.5 & 66.5\\
Student \cite{bergmann2020uninformed}	 				& 85.7 & 91.8 & 86.5 & 91.6 & 69.5 & 81.9 & 93.7 & 81.9 & 89.5 & 93.5 & 92.8 & \textbf{91.2} & 86.3 & 70.1 & 72.5 & 93.3\\
SPADE \cite{cohen2020transformerbased} 				& 91.7 & 95.5 & 90.9 & 93.7 & 94.7 & 86.7 & \textbf{95.4} & 97.2 & \textbf{94.4} & \textbf{94.6} & 96.0 & 75.6 & \textbf{93.5} & \textbf{87.4} & 87.4 & 92.6 \\
PaDiM \cite{defard2020padim}									& 92.1 & 94.8 & 88.8 & 93.5 & 96.2 & 94.6 & 92.6 & 97.8 & 85.6 & 92.7 & 94.4 & 86.0 & 93.1 & 84.5 & \textbf{91.1} & 95.9 \\
\midrule
PatchCore$-25$ & 93.4 & \textbf{96.2} & 92.5 & \textbf{95.5} & \textbf{96.6} & 96.0 & 93.8 &\textbf{ 98.9} & 91.4 & 93.2 & \textbf{97.9} & 87.3 & 91.5 & 83.7 & 89.4 & \textbf{97.1} \\
PatchCore$-10$ & \blue{\textbf{93.5}} & 96.1 & \textbf{92.6} & \textbf{95.5} & \textbf{96.6} & 95.9 & 93.9 & \textbf{98.9} & 91.3 & 94.1 & \textbf{97.9} & 87.4 & 91.4 & 83.5 & 89.6 & \textbf{97.1 }\\
PatchCore$-1$   & 93.1 & 95.9 & 91.6 & \textbf{95.5} & 96.5 & \textbf{96.1} & 93.8 & \textbf{98.9} & 91.2 & 92.9 & 97.1 & 88.3 & 90.2 & 81.2 & 89.5 & 97.0 \\
\bottomrule
\end{tabular}}
\label{tab:appendix_mvtec_sota_seg_pro}
\end{table*}

\begin{table*}[t]
\centering
\caption{Anomaly Detection and Localization Performance (AUROC) on MVTec AD \cite{mvtec} with PatchCore$-1$ using larger images ($280\times 280)$ and a WideResNet101 backbone.}
\resizebox{\textwidth}{!}{
\begin{tabular}{|l||c||c|c|c|c|c|c|c|c|c|c|c|c|c|c|c|}
\toprule
$\downarrow$ Metric \textbackslash Dataset $\rightarrow$& \textbf{Avg} & Bottle & Cable & Capsule & Carpet & Grid & Hazeln. & Leather & Metal Nut & Pill & Screw & Tile & Toothb. & Trans. & Wood & Zipper\\
\midrule
\multicolumn{16}{c}{PatchCore$-1$, Hierarchies (2, 3), Imagesize 280}\\
\midrule
AUROC     & 99.4 & 100 & 99.6 & 98.2 & 98.4 & 99.8 & 100 & 100 & 100 & 97.2 & 98.9 & 98.9 & 100 & 100 & 99.5 & 99.9 \\
pwAUROC   & 98.2 & 98.6 & 98.4 & 99.1 & 98.7 & 98.7 & 98.8 & 99.3 & 98.8 & 97.8 & 99.3 & 96.1 & 98.8 & 96.4 & 95.1 & 98.9 \\
PRO       & 94.4 & 96.6 & 93.8 & 96.0 & 97.4 & 96.8 & 91.2 & 99.1 & 94.8 & 94.0 & 97.5 & 89.5 & 95.5 & 84.8 & 91.7 & 97.8 \\
\midrule
\multicolumn{16}{c}{PatchCore$-1$, Hierarchies (1, 2, 3), Imagesize 280}\\
\midrule
AUROC     & 99.2 & 100 & 99.7 & 98.1 & 98.2 & 98.3 & 100 & 100 & 100 & 97.1 & 99.0 & 98.9 & 98.9 & 99.7 & 99.9 & 99.7 \\
pwAUROC   & 98.4 & 98.6 & 98.7 & 99.1 & 98.7 & 98.8 & 98.8 & 99.3 & 99.0 & 98.6 & 99.5 & 96.3 & 98.9 & 97.1 & 95.2 & 99.0 \\
PRO       & 95.0 & 96.6 & 94.6 & 96.3 & 97.5 & 97.0 & 91.5 & 99.1 & 95.4 & 96.0 & 98.1 & 90.0 & 95.8 & 85.9 & 92.0 & 98.0 \\
\bottomrule
\end{tabular}}
\label{tab:appendix_mvtec_detailed_extended}
\end{table*}

\begin{figure*}
\centering
\begin{subfigure}[b]{0.48\textwidth}
\centering
\includegraphics[width=1\textwidth]{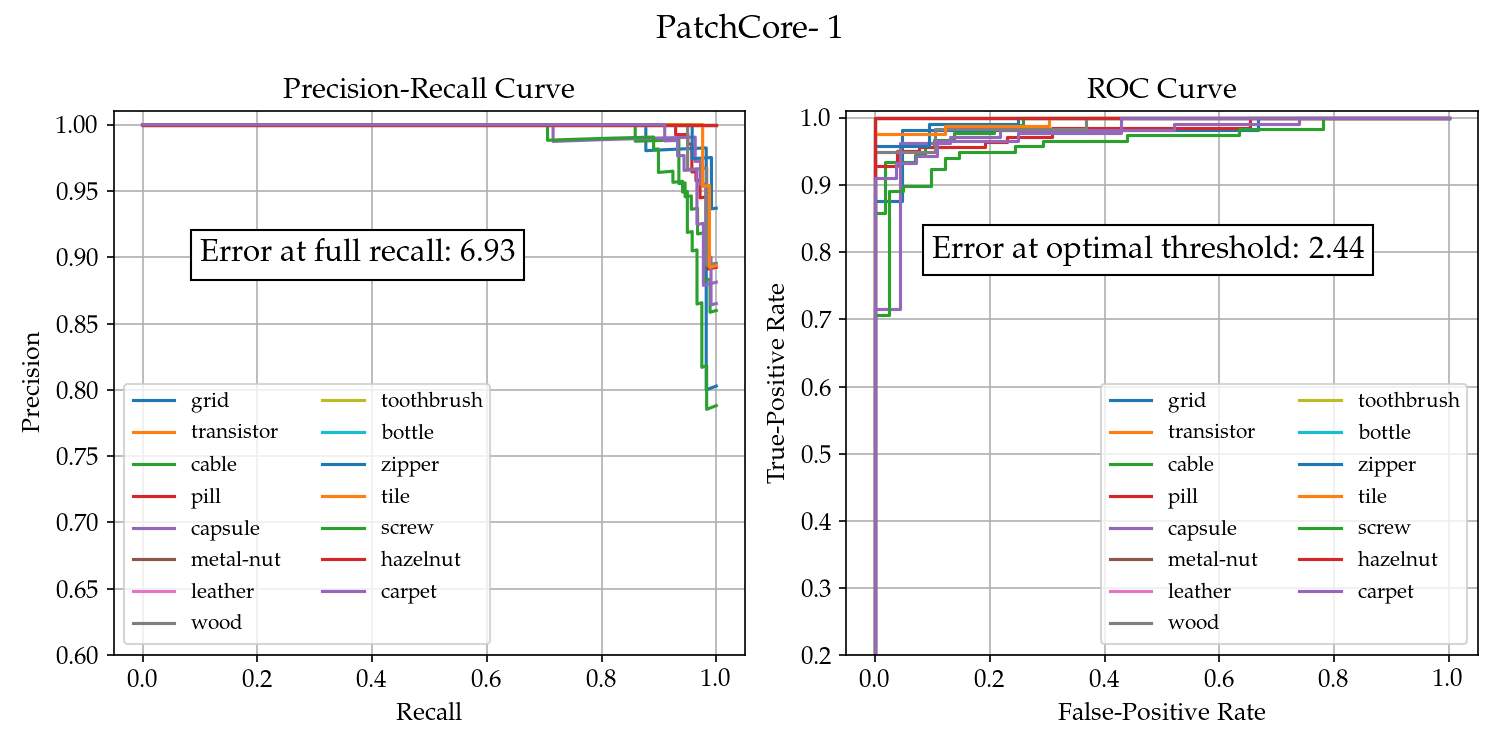}
\end{subfigure}
\begin{subfigure}[b]{0.48\textwidth}
\centering
\includegraphics[width=1\textwidth]{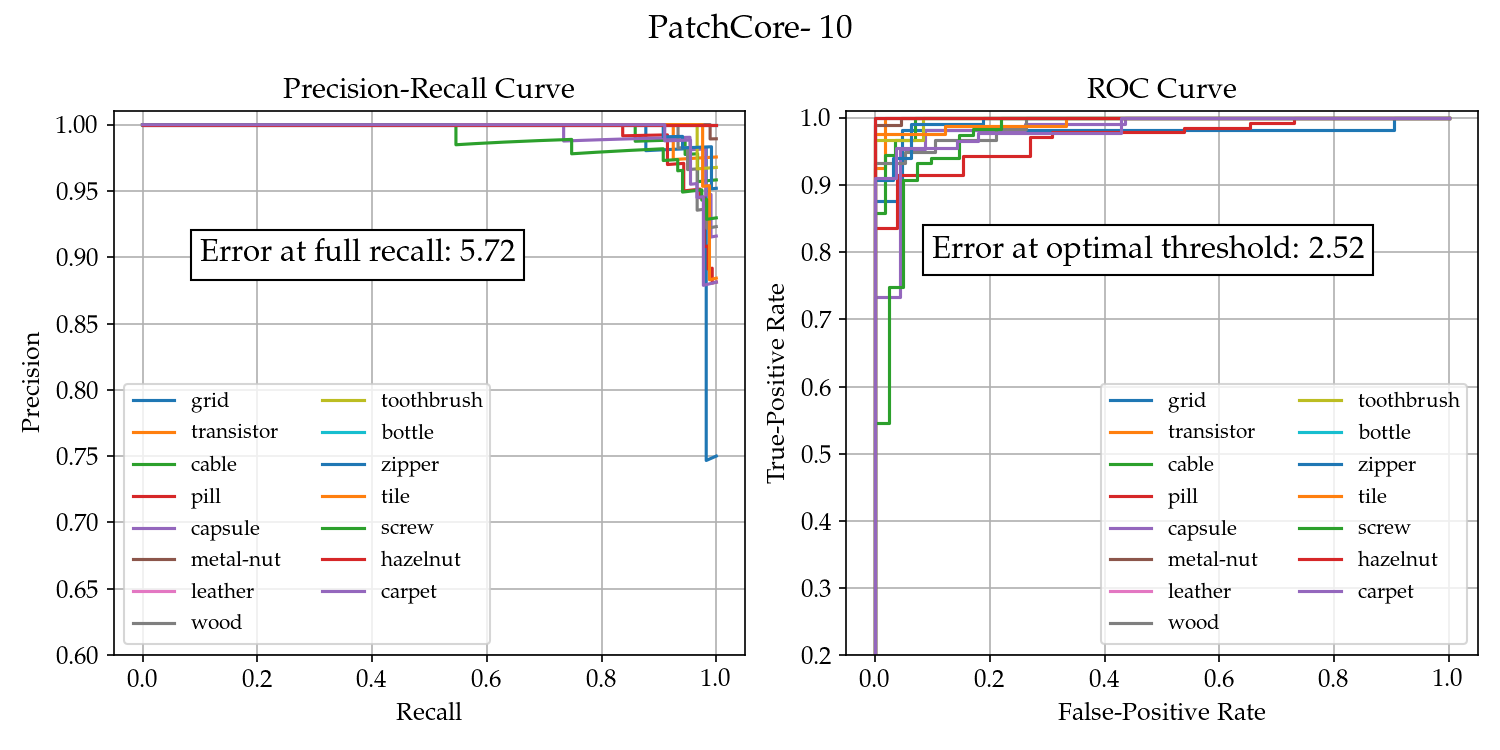}
\end{subfigure}
\centering
\begin{subfigure}[b]{0.48\textwidth}
\centering
\includegraphics[width=1\textwidth]{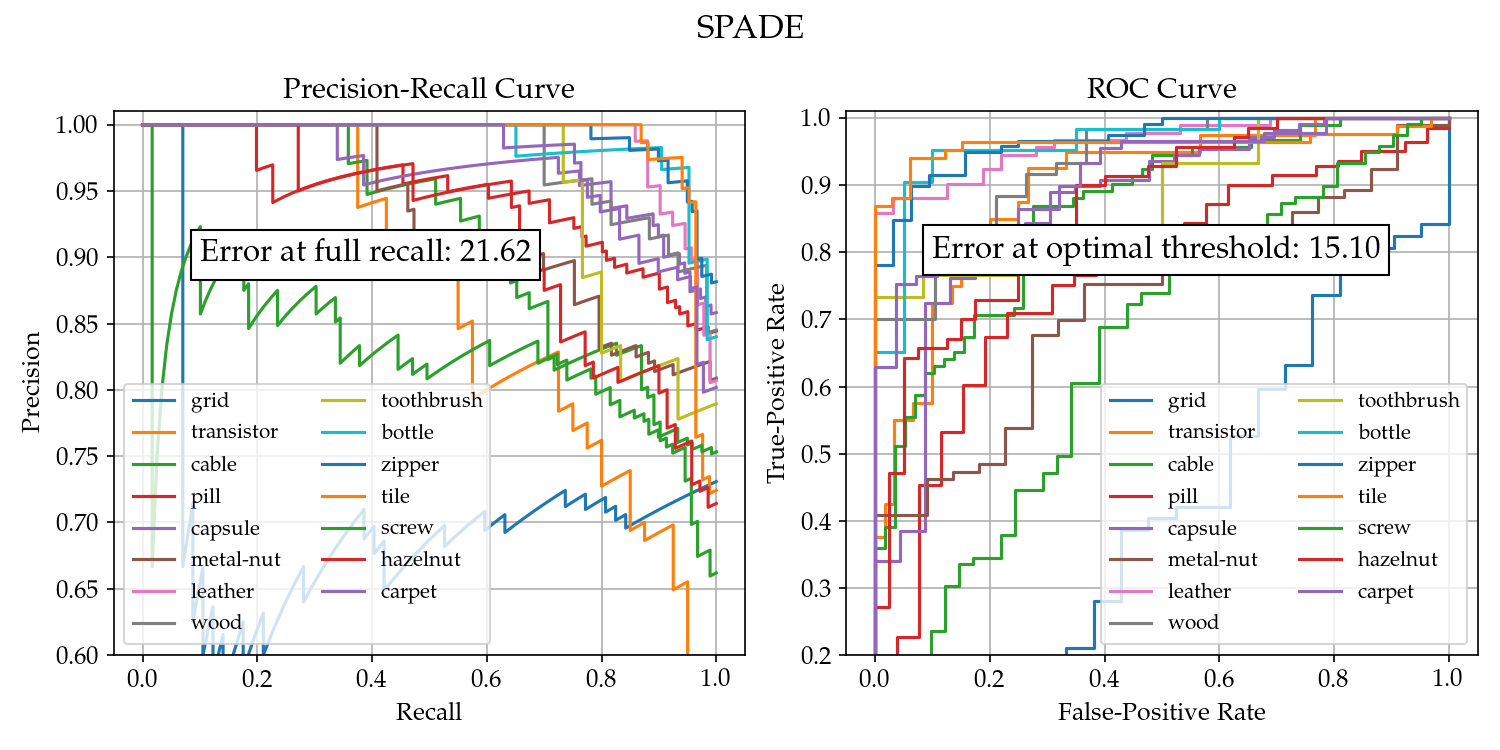}
\end{subfigure}
\begin{subfigure}[b]{0.48\textwidth}
\centering
\includegraphics[width=1\textwidth]{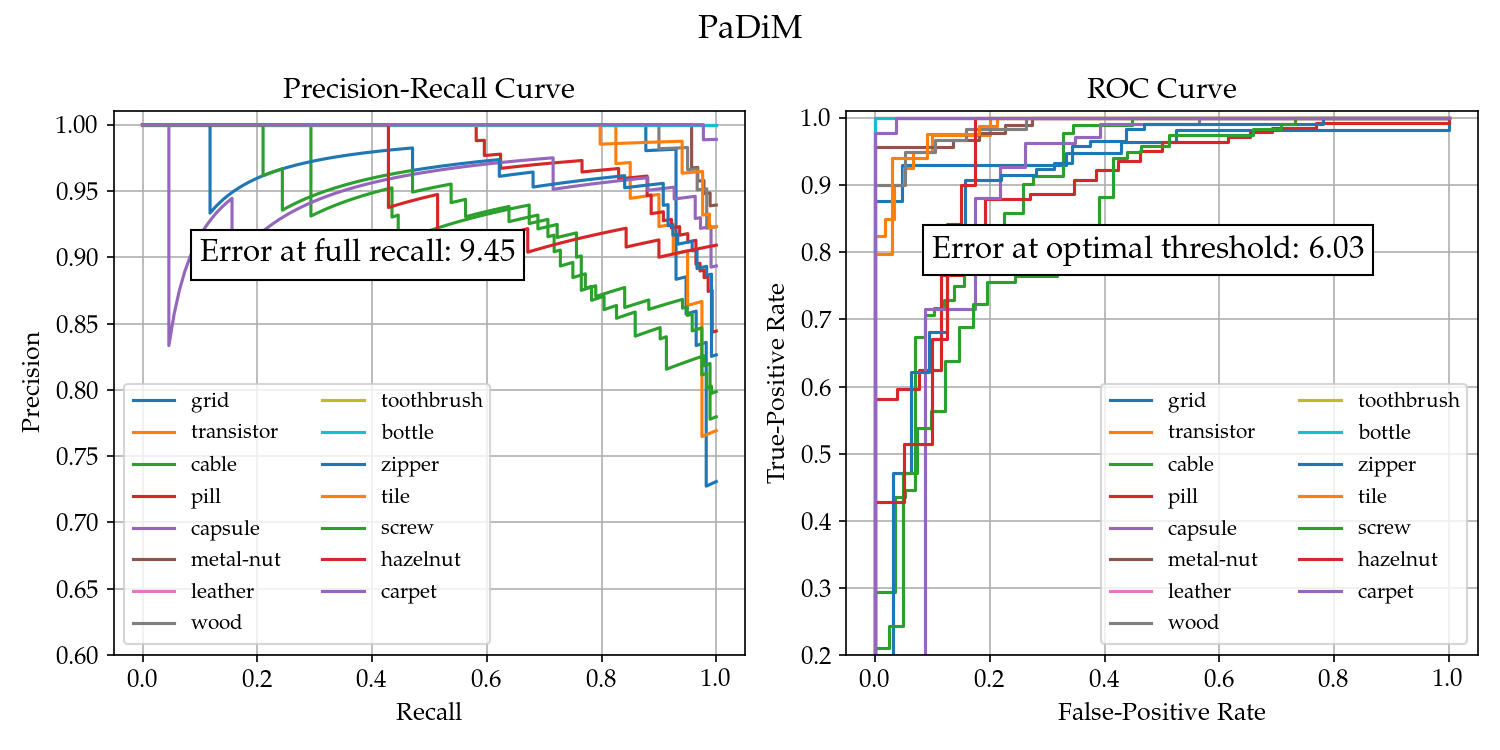}
\end{subfigure}
\caption{Precision-Recall curves (left) and ROC curves (right) for \PatchCore, variants and comparable methods SPADE \cite{cohen2020transformerbased} and PaDiM \cite{defard2020padim}. Different colors in the lines correspond to difference MVTec classes.}
\label{fig:curves}
\end{figure*}

\begin{table*}[h]
\centering
\caption{Low-Shot Anomaly Detection Performance on MVTec \cite{mvtec}, as measured on AUROC.}
\resizebox{0.8\textwidth}{!}{
\begin{tabular}{|c||c|c|c|c|c|c|c|}
\toprule
$\downarrow$ Method \textbackslash Shots $\rightarrow$& 1 & 2 & 5 & 10 & 16 & 20 & 50\\
\midrule
Retained $\%$ & 0.4 & 0.8 & 2.1 & 4.1 & 6.6 & 8.3 & 21\\
\midrule
\multicolumn{8}{c}{\textsc{Image-level AUROC}}\\
\midrule
SPADE & 71.6 $\pm$ 0.7& 73.4 $\pm$ 1.3& 75.2 $\pm$ 1.5& 77.5 $\pm$ 1.1& 78.9 $\pm$ 0.9 & 79.6 $\pm$ 0.8 & 81.1 $\pm$ 0.4\\
PaDiM & 76.1 $\pm$ 0.4 & 78.9 $\pm$ 0.6 & 81.0 $\pm$ 0.2 & 83.2 $\pm$ 0.7 & 85.5 $\pm$ 0.6 & 86.5 $\pm$ 0.3 & 90.1 $\pm$ 0.3\\
DifferNet & - & - & - & - & 87.3 & - & - \\
\hline
PatchCore-10 & 83.4 $\pm$ 0.6& 86.4 $\pm$ 0.9& 90.8 $\pm$ 0.8& 93.6 $\pm$ 0.6 & 95.4 $\pm$ 0.7 & 95.8 $\pm$ 0.6 & 97.5 $\pm$ 0.3\\
PatchCore-25 & 84.1 $\pm$ 0.7& 87.2 $\pm$ 1.0& 91.0 $\pm$ 0.9& 93.8 $\pm$ 0.5 & 95.5 $\pm$ 0.6 & 95.9 $\pm$ 0.6 & 97.7 $\pm$ 0.4\\
\midrule
\multicolumn{8}{c}{\textsc{Pixel-level AUROC}}\\
\midrule
SPADE & 91.9 $\pm$ 0.3 & 93.1 $\pm$ 0.2 & 94.5 $\pm$ 0.1 & 95.4 $\pm$ 0.1 & 95.7 $\pm$ 0.2 & 95.7 $\pm$ 0.2 & 96.2 $\pm$ 0.0\\
PaDiM & 88.2 $\pm$ 0.3 & 90.5 $\pm$ 0.2 & 92.5 $\pm$ 0.1 & 93.9 $\pm$ 0.1 & 94.8 $\pm$ 0.1 & 95.1 $\pm$ 0.1 & 96.3 $\pm$ 0.0\\
\hline
PatchCore-10 & 92.0 $\pm$ 0.2 & 93.1 $\pm$ 0.2 & 94.8 $\pm$ 0.1 & 96.2 $\pm$ 0.1 & 96.8 $\pm$ 0.3 & 96.9 $\pm$ 0.3 & 97.8 $\pm$ 0.0\\
PatchCore-25 & 92.4 $\pm$ 0.3 & 93.3 $\pm$ 0.2 & 94.8 $\pm$ 0.1 & 96.1 $\pm$ 0.1 & 96.8 $\pm$ 0.3 & 96.9 $\pm$ 0.3 & 97.7 $\pm$ 0.0\\
\midrule
\multicolumn{8}{c}{\textsc{PRO Metric}}\\
\midrule
SPADE & 83.5 $\pm$ 0.4 & 85.8 $\pm$ 0.1 & 88.3 $\pm$ 0.2 & 89.6 $\pm$ 0.1 & 90.1 $\pm$ 0.2 & 90.1 $\pm$ 0.3 & 90.8 $\pm$ 0.1\\
PaDiM & 72.4 $\pm$ 1.2 & 77.8 $\pm$ 0.7 & 82.7 $\pm$ 0.2 & 85.9 $\pm$ 0.2 & 87.5 $\pm$ 0.2 & 88.2 $\pm$ 0.2 & 90.4 $\pm$ 0.1\\
\hline
PatchCore-10 & 82.4 $\pm$ 0.3 & 85.1 $\pm$ 0.3 & 88.7 $\pm$ 0.2 & 90.9 $\pm$ 0.1 & 91.8 $\pm$ 0.2 & 92.0 $\pm$ 0.2 & 93.0 $\pm$ 0.1\\
PatchCore-25 & 83.7 $\pm$ 0.5 & 86.0 $\pm$ 0.3 & 88.8 $\pm$ 0.2 & 90.9 $\pm$ 0.1 & 91.7 $\pm$ 0.1 & 91.9 $\pm$ 0.2 & 92.8 $\pm$ 0.0\\
\bottomrule
\end{tabular}}
\label{tab:lowshot_mvtec}
\end{table*}

\begin{table*}[h]
\centering
\caption{Anomaly Detection Performance on MVTec \cite{mvtec}, as measured on AUROC.}
\resizebox{0.45\textwidth}{!}{
\begin{tabular}{|l||c|c|c|c|}
\toprule
$\downarrow$ Backbone & $\%$ of $\mathcal{M}$ & Img. AUROC & Pw. AUROC & PRO \\
\midrule
ResNet50 \cite{He2015resnet}    & 10 & 99.0 & 98.1 & 93.3 \\
											      & 1 & 98.7 & 97.8 & 93.3 \\
\hline											   											 
WideResNet50 \cite{wideresnet} & 10 & 98.9& 98.1 & \textbf{93.5} \\
 & 1 & 99.0 & 98.0 & 93.1 \\
\hline
ResNet101 \cite{He2015resnet}  & 10 & 98.6 & 97.9 & 92.5 \\
 & 1 & 98.7 & 97.8 & 92.2 \\
\hline
WideResNet101 \cite{wideresnet} & 10 & 99.1 & \textbf{98.2} & 93.4 \\
 & 1 & 99.0 & 98.1 & 93.0 \\ 
\hline
ResNeXt101 \cite{resnext} & 10 & 98.9 & 98.0 & 92.8 \\
 & 1 & 98.7 & 97.8 & 92.6 \\
\bottomrule
\end{tabular}}
\label{tab:backbones_mvtec}
\end{table*}


 \begin{figure*}
\centering
\includegraphics[width=0.9\textwidth]{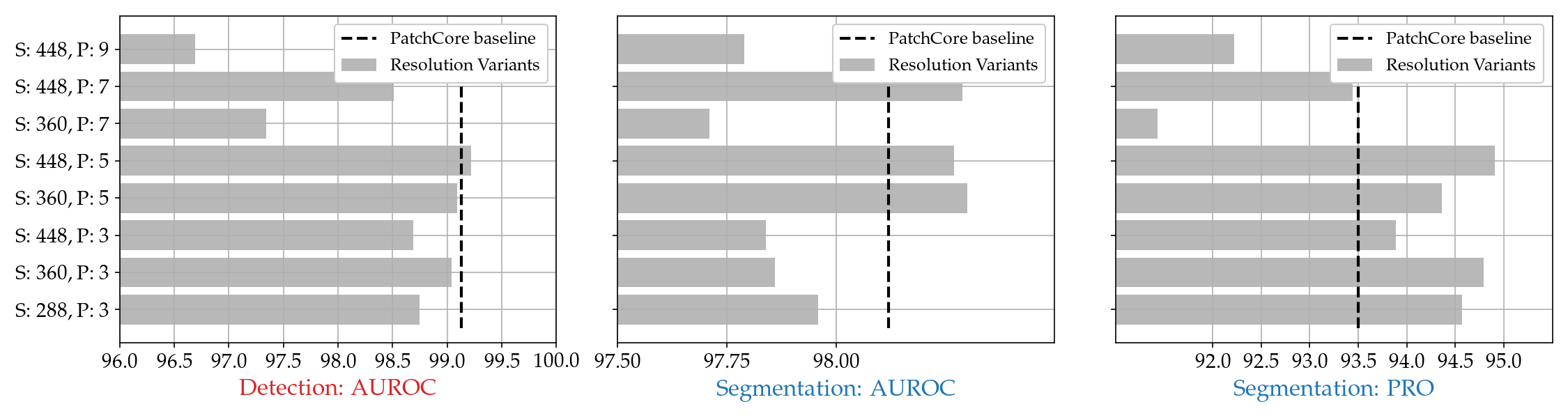}
\caption{Influence of image size (S) and neighbourhood size (P) on \textit{PatchCore} performance. The \PatchCore baseline with default values is included for reference.}
\label{fig:resolution}
\end{figure*}

\begin{figure*}
\centering
\includegraphics[width=1\textwidth]{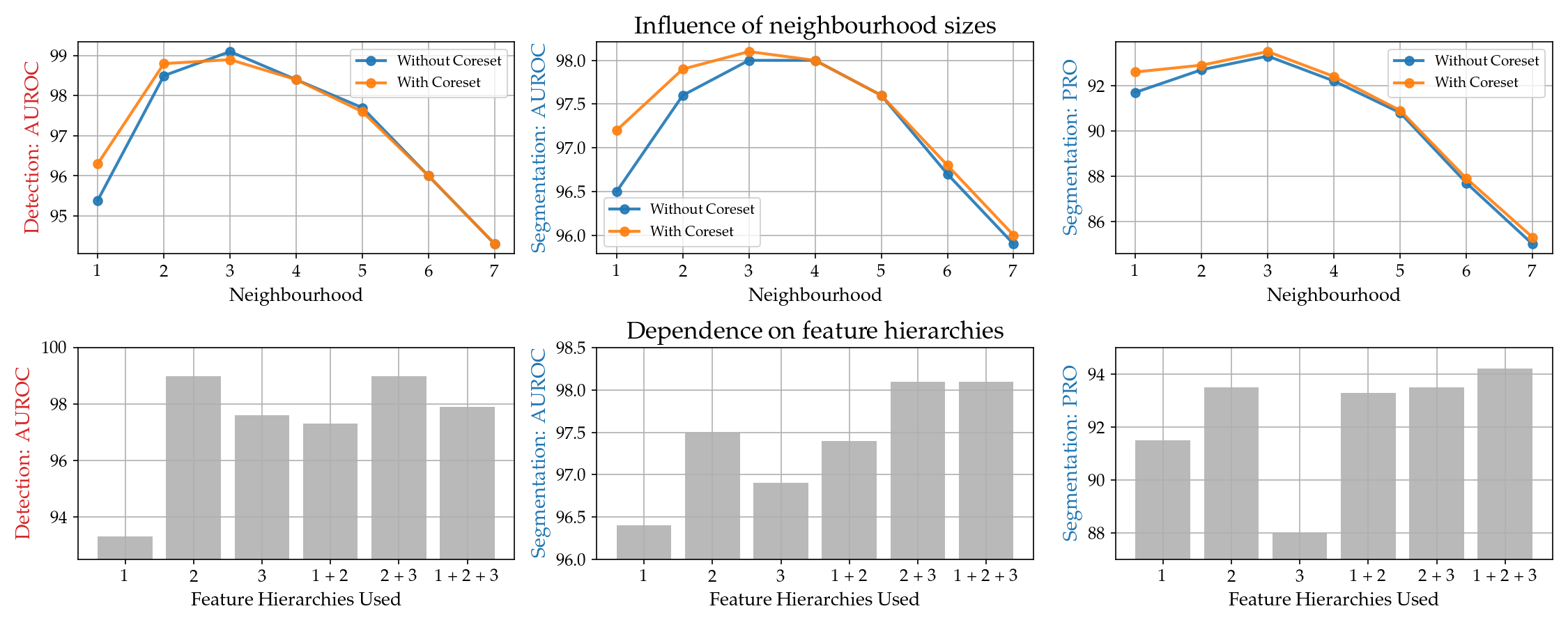}
\caption{Influence of local awareness and network feature depths on anomaly detection performance.}
\label{fig:method_ablation_full}
\end{figure*}

\begin{figure*}
\centering
\includegraphics[width=1\textwidth]{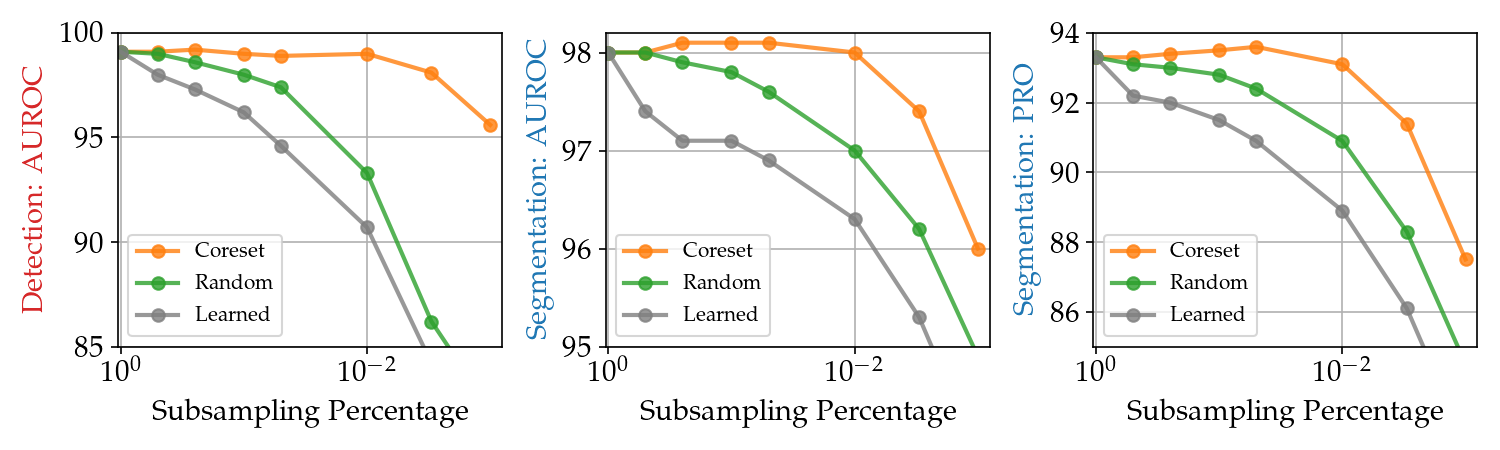}
\caption{Performance retention for different subsamplers.}
\label{fig:subsampling_full}
\end{figure*}

For completeness we repeat the Figures 4 and 5 from the main paper with included PRO score results in
\ref{fig:method_ablation_full} and \ref{fig:subsampling_full}.

\end{document}